\documentclass{article}

\usepackage{microtype}
\usepackage{graphicx}
\usepackage{subcaption}
\usepackage{booktabs} 

\usepackage{hyperref}

\usepackage[preprint]{icml2026}

\usepackage{amsmath}
\usepackage{amssymb}
\usepackage{mathtools}
\usepackage{amsthm}

\usepackage[capitalize,noabbrev]{cleveref}

\theoremstyle{plain}

\theoremstyle{definition}

\theoremstyle{remark}

\usepackage[textsize=tiny]{todonotes}

\icmltitlerunning{Learning Interpretable PDE Representations for Generative Reconstructions with Structured Sparsity}

\begin{document}

\twocolumn[
  \icmltitle{Learning Interpretable PDE Representations for Generative Reconstructions with Structured Sparsity}



  \icmlsetsymbol{equal}{*}

  \begin{icmlauthorlist}
    \icmlauthor{Valerie Tsao}{duke}
    \icmlauthor{Nathaniel Chaney}{duke}
    \icmlauthor{Manolis Veveakis}{duke}
  \end{icmlauthorlist}

  \icmlaffiliation{duke}{Department of Civil \& Environmental Engineering, Duke University, NC, USA}

  \icmlcorrespondingauthor{Valerie Tsao}{valerie.tsao@duke.edu}
  \icmlkeywords{Machine Learning, ICML, Latent Diffusion}

  \vskip 0.3in
]



\printAffiliationsAndNotice{}  

\begin{abstract}
  Scientific measurements are often bottlenecked by suboptimal conditions, whether that be noise, incomplete spatial coverage, or limited resolution, rendering accurate field reconstruction a difficult task. We introduce LatentPDE, a latent diffusion framework designed to simultaneously resolve sparse-observation reconstruction and super-resolution. While existing physics-guided diffusion models typically rely on soft loss penalties or uninterpretable representations, our approach enforces physical compliance by constructing an inherently interpretable latent space. Specifically, we parameterize the latent variables directly as the coefficients and source terms of an assumed governing PDE. In doing so, LatentPDE is able to reliably reconstruct dynamics across highly disparate and structured data gaps. Empirical results on diverse configurations demonstrate that our model achieves high-fidelity recovery at any desired resolution while also tracking the underlying predictive uncertainty. 
\end{abstract}

\section{Introduction}
Physical systems in real-world settings are canonically difficult to emulate and monitor. Observation stations are sparse, and measurements are corrupted and typically limited in resolution by technological or financial barriers. When direct measurements are deficient and unable to fully define a system, inverse modeling bridges these observational gaps by relying on governing partial differential equations (PDEs) as structural constraints.


Existing inverse methods, however, fail to bridge these gaps consistently. Classical approaches scale poorly to high-dimensional states and lack the ability to quantify uncertainty. Recent innovations in  operator-based \cite{kovachki2023neural, huang2024operator} and physics-informed neural networks have made progress on this front  \cite{baldan2023electromagnetic, zheng2024hompinns}, but generally under idealized assumptions of clean, full-coverage data fields at desired resolutions, and at most only address one of these challenges at a time. Furthermore, these methods tend to treat governing physics as a soft regularizer meant to \textit{guide} rather than \textit{enforce} constraints, allowing models to still violate physics at inference time. This can lead to oversmoothed reconstructions that miss important system dynamics. In high-stakes applications like localized extreme weather forecasting and medical imaging \cite{kashinath2021physics, gao2021super}, decision-makers cannot make informed assessments on such ambiguous outputs. They require both physically aligned reconstructions and calibrated confidence intervals to safely determine risk levels and mitigate for worst-case scenarios.

\begin{figure*}[t]
    \centering
    \includegraphics[width=.9\textwidth]{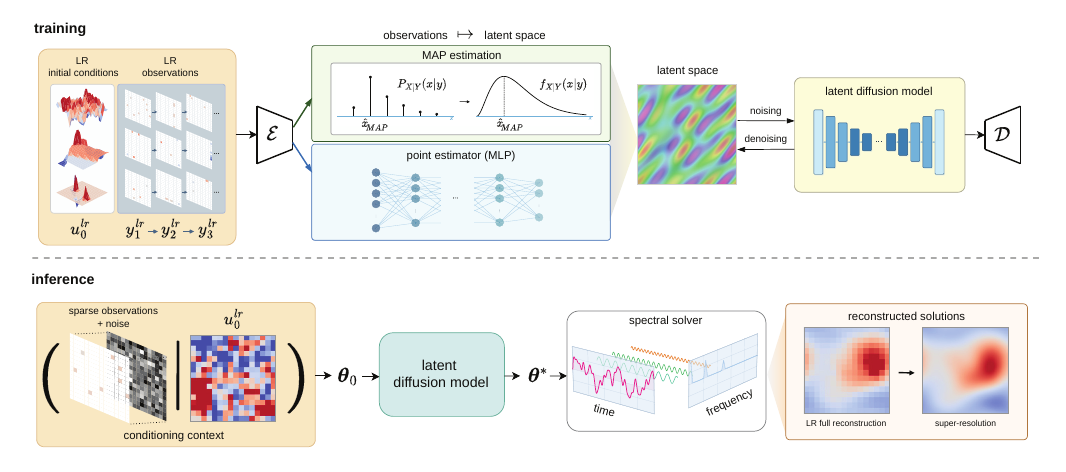}
    \caption{Schematic of LatentPDE framework. Sparse, noised, low-resolution observations and initial conditions ($u_0^{lr}$) are encoded (via MAP or MLP) into a latent space of unknown PDE coefficients. A noising-denoising algorithm refines these parameters before a spectral solver decodes them, recovering full and high-resolution fields that strictly enforce governing physics.}
    \label{fig:framework}
\end{figure*}

We propose a latent diffusion framework that trains directly on sparse fields while explicitly embedding governing laws via an interpretable latent space. Rather than applying a physics penalty to our loss term, our core insight is to embed the governing laws directly into the architecture by using PDE coefficients as our latent representation. By decoding this latent space via a differentiable spectral solver, the framework imposes physics as a structural constraint on the solution manifold. In doing so, we are able to accommodate highly structured missingness rather than merely handling uniform random sparsity. This hard-enforcement in the spectral domain achieves super-resolution at inference and provides the inductive bias needed to solve traditionally ill-posed sparse reconstructions. To our knowledge, LatentPDE is the first to directly couple the denoising framework of diffusion models with the coefficient space, thereby guaranteeing a physically-plausible reconstruction while providing intrinsic uncertainty quantification on noisy, spatially incomplete data at sub-target resolutions. 




We empirically validate LatentPDE across three separate classes of PDEs: advection--diffusion (parabolic), Klein--Gordon (hyperbolic), and Helmholtz (elliptical). The framework is benchmarked against a wide set of challenging initial conditions and diverse source terms as defined by a 2D Fourier expansion. Even when evaluated against out-of-distribution (OOD) masks, our framework surpasses other state-of-the-art methods on full reconstructions. Simultaneously, it characterizes a posterior distribution over the governing parameters to yield confidence estimates. LatentPDE establishes a bounded generative framework for inferring dynamic fields across space and time, making it suitable for any environment requiring safe generations, from structural health monitoring to climate resilience.

\section{Background}
\subsection{PDE Solvers and the Inverse Problem}
PDEs describe the mathematical relationships governing complex physical systems. With known governing parameters and initial/boundary conditions, spatiotemporal states of the system can be inferred using forward solvers. For many PDEs, however, analytic solutions remain largely intractable. As such, many approaches must make use of discretization schemes such as Finite Element Method (FEM) \cite{felippa2004introduction} and Finite Difference Method (FDM) \cite{thomas2013numerical} to approximate these solutions numerically.

Among these schemes, spectral methods provide an efficient mechanism by decomposing numerical solutions into a finite expansion of global basis functions. Their core intuition relies on transforming differentiation to a simple operation on the associated coefficients, achieving high accuracy for smooth solutions with far fewer degrees of freedom than local discretization methods \cite{shen2011spectral}.  Still, classical forward solvers require prior knowledge of the governing coefficients. In practice, we only have access to indirect and sparse observations, necessitating the solution to the \textit{inverse} problem: inferring missing parameters from observations. Because this inverse mapping is ill-posed and under-determined, spectral representations have been increasingly coupled with probabilistic frameworks to infer incomplete dynamics and forecast continuous fields \cite{lv2025hybrid, arnst2013hybrid}.

\subsection{Limitations of Existing Inverse Solvers}
One of the earliest methods to address the inverse problem was Tikhonov regularization \cite{tikhonov1943stability}. Subsequent approaches came in the form of variational Bayesian inference methods, which recast posterior inference as the optimization of a surrogate distribution within a chosen variational family \cite{Stuart_2010}. In practice, two classes of methods have emerged for data assimilation (DA) in physical systems: variational approaches such as 4D-Var \cite{courtier4dvar}, which optimize over a time window via adjoint methods but yield no posterior distribution, and ensemble methods such as the EnKF \cite{evensen1994sequential, iglesias2013ensemble}, which propagate uncertainty through an ensemble-based linearization but struggle with non-Gaussian, high-dimensional state spaces. However, the inability of Bayesian approaches to scale or handle structured missingness without strong parametric assumptions can make them vastly restrictive when compared to the flexibility of data-driven approaches.

More recently, data-driven operator learning methods such as physics-informed neural networks (PINNs) \cite{RAISSI2019686}, Fourier Neural Operators (FNO) \cite{li2020fourier}, and DeepONet \cite{lu2019deeponet} have been used to learn complex prior structures. However, these architectures are fundamentally deterministic point estimators with no mechanism for representing solution uncertainty. On the other hand, amortized approaches such as simulation-based inference \cite{deistler2025simulation} and neural posterior estimation \cite{ward2022robust} offer scalable posterior approximation, but do not exploit the structural regularity of the PDE solution manifold.

\subsection{Generative Models and Latent Structures}
Diffusion models provide a solution to this need for uncertainty quantification. In many ways, they can be thought of as hierarchical VAEs \cite{kingma2013auto} which work by gradually adding noise to data samples in a forward Markov process. Once completely noised, a deep neural network learns to denoise the corrupted samples such that they can arbitrarily generate new samples consistent with the input distribution \cite{sohl2015deep, ho2020denoising}. 

To reduce the computational overhead of standard diffusion models, recent approaches have shifted to latent formulations \cite{rombach2022high}. Yet, the question of imposing structure on learned latent spaces remains an open area of investigation under the field of disentanglement \cite{kim2018disentangling, hinz2018inferencing, wang2024disentangled}, which seeks to find representations where individual dimensions correspond to interpretable generative factors. While approaches to impose physical guidance or simplify the problem in a compressed observation space exist, the former is typically not strictly enforced \cite{bastek2024physics}, and the latter is most often uninterpretable \cite{chefer2023hidden}, making the latent-to-observable mapping a black-box mechanism that cannot verify if solutions are physically consistent.


\section{Problem Setup}
We study the joint problem of \emph{sparse-observation reconstruction} and \emph{super-resolution} for fields governed by linear PDEs on the periodic unit square $\Omega = [0,1)^2$. At inference time, the model receives only a sparse, noisy snapshot of a coarse output field $y$ together with a low-resolution initial condition $u_0^{\text{lr}}$. The goal is to find a physically-consistent latent representation that (i) recovers the governing coefficients agnostic of noise or sparsity present in the observations and (ii) allows for seamless upscaling to any desired resolution.
\subsection{Observation Model}
We work with a high-resolution grid of size $H_\mathrm{hr} \times W_\mathrm{hr}$ and a low-resolution grid obtained by average pooling with integer factor $p$. Let $u_\mathrm{out}^\mathrm{hr}$ denote the high-resolution output field (the terminal state $u(\cdot, T)$ for evolutionary PDEs). Sparse observations are drawn as
\begin{equation}\label{eq:obs_model}
    y = M \odot D_p(u_\mathrm{out}^\mathrm{hr}) + \sigma_\mathrm{obs}\, M \odot \varepsilon, \quad \varepsilon \sim \mathcal{N}(0, I),
\end{equation}
where $M \in \{0,1\}^{H_\mathrm{lr} \times W_\mathrm{lr}}$ is a binary observation mask and $D_p$ denotes average pooling.

\subsection{Latent Physical Parameterization}
Each candidate reconstruction is represented by a raw latent
\begin{equation}
    \mathbf{z} = [r_1, r_2, r_3,\, \mathbf{q}] \in \mathbb{R}^{3 + 4M_\mathrm{force}^2},
\end{equation}
with three components. First, $(r_1, r_2, r_3)$ are mapped to physical coefficients $\boldsymbol{\theta}$ via interval-constrained sigmoids, so the same latent geometry accommodates all families (Appendix~\ref{sec:appendix_latent}). For our forcing term, we take inspiration from the spectral latent representations introduced by FNOs \cite{li2020fourier}. Here, $\mathbf{q}$ stores a compact low-mode Fourier representation of the steady forcing field. A further discussion of this term is presented in Section~\ref{subsec:forcing}. During inference, raw latents are normalized coordinate-wise using dataset statistics.

\subsection{PDE Families and Spectral Solver}
All PDE families in our experiments are linear, constant-coefficient, and periodic, so each spatial Fourier mode evolves independently. This yields a unified spectral transfer representation
\begin{equation}
    \widehat{u}_\mathrm{out}(\mathbf{k}) = G_\mathfrak{f}(\mathbf{k}, T;\, \boldsymbol{\theta})\, \widehat{u}_0(\mathbf{k}) + H_\mathfrak{f}(\mathbf{k}, T;\, \boldsymbol{\theta})\, \widehat{f}(\mathbf{k}),
    \label{eq:spectral_transfer}
\end{equation}
where $\mathfrak{f}$ indexes the PDE family, $\boldsymbol{\theta}$ are its physical parameters, and $G_\mathfrak{f}, H_\mathfrak{f}$ are family-specific transfer multipliers. Equation~\eqref{eq:spectral_transfer} is what allows us to use a single decoder to span multiple PDE types: recovering $u_\mathrm{out}$ at any resolution reduces to inverting an FFT once the latent $(\boldsymbol{\theta}, \widehat{f})$ is known. The full forms of $G_\mathfrak{f}$ and $H_\mathfrak{f}$ for each particular family can be found in Appendix~\ref{sec:appendix_problem_setup}.

\section{Methods}
\label{sec:methods}
Our framework introduces an approach to recovering a suitable latent variable $\mathbf{z}$ that, once decoded, corresponds to a physically and spectrally-consistent reconstruction of each observation $y$. Concretely, inference proceeds in three stages: an encoder $\mathcal{E}$ maps sparse observations (conditioned on the low-resolution initial condition) to an initial latent estimate $\mathbf{z}_0$. Then a denoising process refines $\mathbf{z}_0$ toward a posterior sample $\mathbf{z}^*$. Finally, the spectral decoder $\mathcal{D}$ maps $\mathbf{z}^*$ to a high-resolution, physics-consistent output field. Schematically:

\begin{equation}\label{eq:schematic}
    y \mid u_0^{\text{lr}}\overset{\mathcal{E}}{\mapsto} \mathbf{z}_0 \xrightarrow{\text{denoising}} \mathbf{z}^* \overset{\mathcal{D}}{\mapsto} u_{\text{out}}^{\text{hr}}.
\end{equation}

Because $\mathcal{D}$ is fixed and differentiable across all methods, each encoder strategy reduces to the same structured inverse problem of finding the $\mathbf{z}$ whose decoded field is consistent with the data under the objective
\begin{equation}
    \mathcal{J}(\mathbf{z};\, y, M, u_0^{\mathrm{lr}}, \mathfrak{f})
    \;=\;
    \frac{\bigl\| M \odot \bigl[\mathcal{D}(\mathbf{z},
    u_0^{\mathrm{lr}};\mathfrak{f}) - y\bigr] \bigr\|_2^2}{\|M\|_1 + \varepsilon}.
    \label{eq:residual}
\end{equation}
We introduce two strategies for the encoder $\mathcal{E}$: direct test-time optimization (\textsc{LatentPDE-MAP}) and amortized prediction (\textsc{LatentPDE-Enc}). We use each as a warm-start initialization for a latent diffusion model that further improves upon the estimate and captures posterior uncertainty. This design cleanly separates the inference mechanism from the PDE solver. Thus, all performance differences among methods are attributable to how the latent is inferred, not to changes in the underlying physics.

\subsection{Conditioning Representation}
\label{subsec:conditioning}

Before any encoding is done, the inference models first condition on sparse observations through a four-channel spatial representation consisting of the masked field \(y\), the binary mask \(M\), a distance-to-observation map \(d(M)\), and a local observation-density map \(\rho(M)\). Stacking these as $\bigl[y,\, M,\, d(M),\, \rho(M)\bigr]$ lets the network distinguish unobserved zeros from true zero measurements and exposes the spatial structure of the sensor pattern, which is critical when test masks differ from those seen during training. A shared CNN sparse encoder processes this four-channel input into a fixed-size embedding $\mathbf{c}_{\mathrm{obs}}$ and a separate small CNN encodes $u_0^{\mathrm{lr}}$ into $\mathbf{c}_{u_0}$.  Full architectural details are shown in Appendix~\ref{sec:appendix_architecture}. Both embeddings are passed to each of the two encoder strategies described below, which differ only in how they use the embeddings to produce a latent estimate.

\subsection{LatentPDE-MAP}
\label{subsec:physics_map}
\emph{Maximum a posteriori} (MAP) estimation recovers the most probable latent by directly minimizing a regularized version of Eq.~\eqref{eq:residual} at test time, with no learned components beyond the decoder itself. For each test sample we solve
\begin{equation}
    \mathbf{z}_{\mathrm{MAP}} 
    = \arg\min_{\mathbf{z}}\;
    \mathcal{J}(\mathbf{z};\, y, M, u_0^{\mathrm{lr}}, \mathfrak{f})
    \;+\; \lambda_{\mathrm{MAP}} \|\mathbf{z}\|_2^2,
    \label{eq:map_obj}
\end{equation}
with $\lambda_{\mathrm{MAP}} = 0.01$. Because the masked residual is non-convex and can harbor local minima under very sparse masks, we run 300 Adam steps from three independent random initializations and retain the restart with the lowest masked residual. MAP requires no training data and is guaranteed to fit the observations, but its per-sample optimization is slower than amortized alternatives. \textsc{LatentPDE-Enc} trades this flexibility for near-instant inference.

\subsection{LatentPDE-Enc}
\label{subsec:encoder}
Rather than optimizing per-sample, the amortized encoder $E_\psi$ predicts the latent in a single forward pass:
\begin{equation}
    \hat{\mathbf{z}}_{\mathrm{enc}} = E_\psi(y, M, u_0^{\mathrm{lr}}),
\end{equation}
where $E_\psi$ is a three-layer MLP that fuses $\mathbf{c}_{\mathrm{obs}}$ and $\mathbf{c}_{u_0}$, and is trained end-to-end through the fixed decoder with
\begin{equation}
    \mathcal{L}_{\mathrm{enc}}
    = \mathcal{J}(\hat{\mathbf{z}}_{\mathrm{enc}};\, y, M, u_0^{\mathrm{lr}},
    \mathfrak{f})
    \;+\; \lambda_{\mathrm{enc}} \|\hat{\mathbf{z}}_{\mathrm{enc}}\|_2^2,
    \label{eq:enc_loss}
\end{equation}
where $\lambda_{\mathrm{enc}} = 0.01$. Training details are given in Appendix~\ref{sec:appendix_architecture}. While amortization yields fast inference, by itself the encoder may generalize poorly to unseen mask patterns during training, hence the need for a diffusion framework to refine this initialization.

\subsection{Latent Diffusion Models}
\label{subsec:latent_diffusion}
We train denoising diffusion models \cite{ho2020denoising} in normalized latent space \cite{rombach2022high} to learn a prior over physically-consistent latents and to provide calibrated uncertainty. The forward (noising) process follows a linear variance schedule over $T = 400$ steps:
\begin{equation}
    q(\mathbf{z}_t \mid \mathbf{z}_0)
    = \mathcal{N}\!\bigl(\sqrt{\bar\alpha_t}\,\mathbf{z}_0,\;
    (1 - \bar\alpha_t) I\bigr).
    \label{eq:forward_diffusion}
\end{equation}
The denoiser $\epsilon_\phi$ takes as input the noisy latent, the conditioning representation $[\mathbf{c}_{\mathrm{obs}}, \mathbf{c}_{u_0}]$, and a sinusoidal timestep embedding. These are injected into a six-block residual MLP via FiLM modulation~\cite{perez2018film}.  We train two models with identical architecture but different target latent distributions: \textsc{LatentPDE-MAP} uses latents $\{\mathbf{z}_{\mathrm{MAP}}^{(i)}\}$ obtained by running the MAP objective on each training sample, while \textsc{LatentPDE-Enc} uses encoder predictions $\{\hat{\mathbf{z}}_{\mathrm{enc}}^{(i)}\}$.

\begin{figure*}[h!]
    \centering
    \includegraphics[width=.85\textwidth]{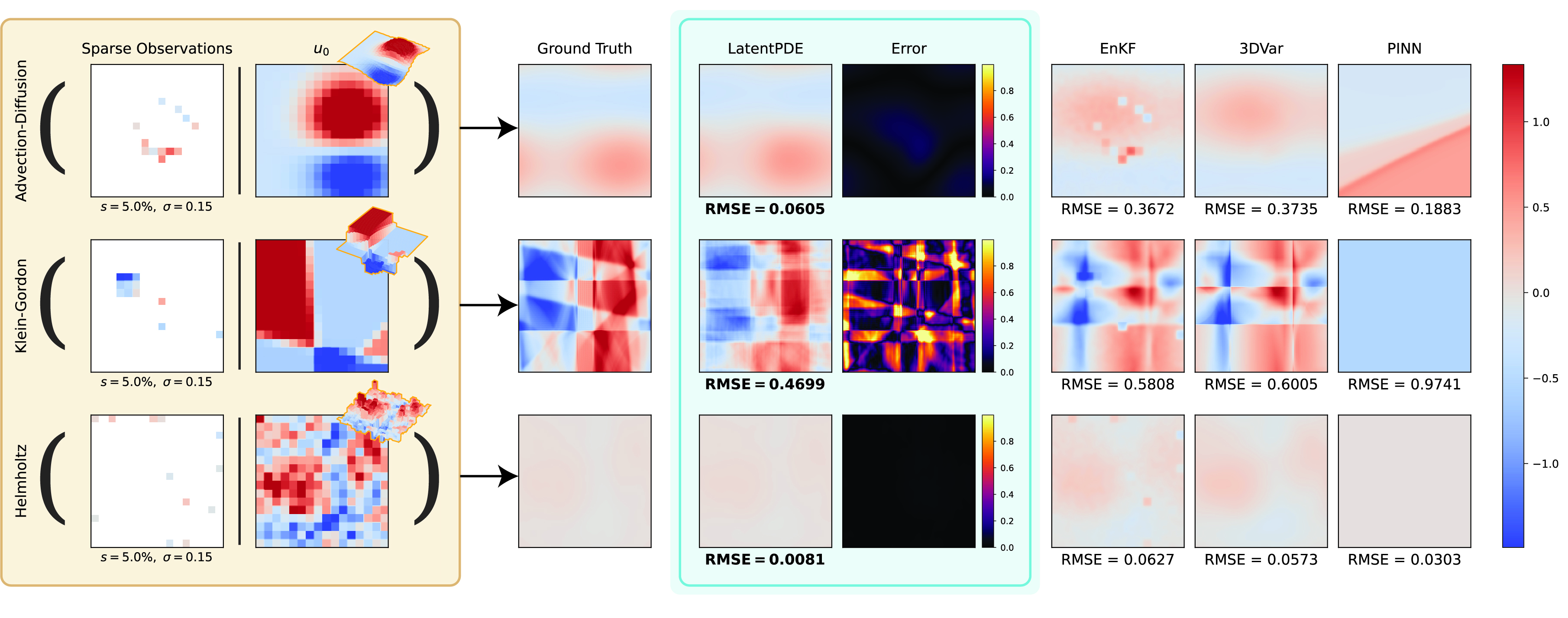}
    \caption{Full-field reconstruction of PDEs from sparse ($s = 5.0\%$) and noisy ($\sigma = 0.15$) observations. \textbf{Note that LatentPDE is conditioned here on the low-resolution initial state ($u_0$) shown, demonstrating simultaneous super-resolution and reconstruction functionality, whereas other baselines require a high-resolution $u_0$ (not pictured).} Despite this, LatentPDE consistently achieves the highest fidelity, reducing RMSE by over 87\% compared to EnKF (Helmholtz) and up to 73\% relative to the next-best performing method (PINNs, Klein--Gordon). By addressing measurement noise and incomplete coverage, LatentPDE successfully reconstructs the full physical solutions without sacrificing complex spatial dynamics.}
    \label{fig:main_results}
\end{figure*}

The training loss augments the standard denoising objective with an observation-consistency term weighted by $\bar\alpha_t$ to emphasize cleaner timesteps:
\begin{align}
    \mathcal{L}_{\mathrm{diff}}
    &= \mathcal{L}_{\mathrm{denoise}}
    + \lambda_{\mathrm{obs}}\,\mathcal{L}_{\mathrm{obs}},
    \label{eq:diff_loss}\\
    \mathcal{L}_{\mathrm{obs}}
    &= \mathbb{E}_t\!\Bigl[
    \bar\alpha_t\;
    \mathcal{J}(\hat{\mathbf{z}}_0;\, y, M, u_0^{\mathrm{lr}},\mathfrak{f})
    \Bigr],
    \label{eq:obs_consistency}
\end{align}
where $\hat{\mathbf{z}}_0$ is the clean-latent prediction derived from the denoiser output. As in the standard denoising objective, the expectation over $t$ is estimated by sampling diffusion timesteps uniformly during minibatch training. This trains the same denoiser to form usable clean-latent estimates across the whole noising schedule rather than at a single selected noise level. $\bar{\alpha}_{t}$  downweights the decoded physical residual at high-noise timesteps, where the clean latent estimate divides by a small $\sqrt{\bar{\alpha}_{t}}$ and is therefore less reliable. Conditioning is then dropped with probability $0.1$ during training for robustness~\cite{ho2022classifierfree}.

\subsection{Posterior Sampling with Physics Guidance}
\label{subsec:posterior_sampling}

At test time the reverse chain begins from a noisy perturbation of the conditioning-specific initialization $\mathbf{z}_{\mathrm{MAP}}$  or $\hat{\mathbf{z}}_{\mathrm{enc}}$. Ensemble diversity is induced by varying the injected noise level $\sigma_{\mathrm{init}} \in [0.3, 0.6]$ across members.

Because the denoiser is only \emph{approximately} conditioned on observations, we apply physics guidance~\cite{chung2022dps} at each reverse step for $t < 0.8\,T$. The predicted clean latent $\hat{\mathbf{z}}_0$ is refined by clipped gradient descent on the masked residual,
\begin{equation}
    \hat{\mathbf{z}}_0 \leftarrow \hat{\mathbf{z}}_0 - \eta_g\,\gamma_g\; \operatorname{clip}\!\bigl( \nabla_{\hat{\mathbf{z}}_0}\mathcal{J}(\hat{\mathbf{z}}_0),\,{-5},{5} \bigr),
    \label{eq:guidance}
\end{equation}
with $\eta_g = 0.10$, $\gamma_g = 80$, and $3$ inner steps per reverse
transition.  After the chain completes, an optional $20$-step Adam refinement minimizes $\mathcal{J}(\mathbf{z}) + \lambda_{\mathrm{ref}}\|\mathbf{z} - \mathbf{z}_{\mathrm{diff}}\|_2^2$ to reduce residual mismatch while staying anchored to the diffusion output.  The full procedure is summarized in Algorithm~\ref{alg:diffusion_sampling}, and an ablation study of inference-time adjustments can be found in Appendix~\ref{sec:appendix_inference_ablation}.

\begin{algorithm}[t]
\caption{Posterior sampling with physics guidance}
\label{alg:diffusion_sampling}
\begin{algorithmic}[1]
\STATE \textbf{Input:} observations $y$, mask $M$, IC $u_0^{\mathrm{lr}}$,
       denoiser $\epsilon_\phi$, PDE family $\mathfrak{f}$
\STATE Compute $\mathbf{z}_{\mathrm{init}}$ via MAP or encoder
\STATE Sample $\mathbf{z}_{T_{\mathrm{init}}}$ as noisy perturbation of
       $\mathbf{z}_{\mathrm{init}}$ at level $\sigma_{\mathrm{init}}$
\FOR{$t = T_{\mathrm{init}}, \ldots, 0$}
    \STATE Predict $\hat\epsilon_t = \epsilon_\phi(\mathbf{z}_t, y, M, u_0^{\mathrm{lr}}, t/T)$; form $\hat{\mathbf{z}}_0$
    \IF{$t < 0.8\,T$}
        \STATE Update $\hat{\mathbf{z}}_0$ via physics guidance (Eq.~\ref{eq:guidance}, 3 steps)
    \ENDIF
    \STATE DDPM reverse step $\rightarrow \mathbf{z}_{t-1}$
\ENDFOR
\STATE \textit{(Optional)} Refine $\mathbf{z}_0$ via Adam on
       $\mathcal{J}(\mathbf{z}_0) + \lambda_{\mathrm{ref}}\|\mathbf{z}_0 - \mathbf{z}_{\mathrm{diff}}\|_2^2$
\STATE \textbf{Return} $\mathcal{D}(\mathbf{z}_0, u_0^{\mathrm{out}})$
\end{algorithmic}
\end{algorithm}

\section{Experimental Results}
All models were trained and all experiments were run on a single NVIDIA RTX A5000 GPU.
\subsection{Data and Mask Generation}
\label{sec:datamask_gen}
We generate synthetic inverse-problem pairs by first sampling a high-resolution initial condition $u_0^{\mathrm{hr}}$, then drawing regime-specific PDE parameters and forcing coefficients, and solving the corresponding PDE with a spectral solver to obtain the high-resolution target $u_{\text{out}}^{\mathrm{hr}}$. To construct the low-resolution counterpart, we average-pool $u_0^{\mathrm{hr}}$ to $u_0^{\mathrm{lr}}$ and solve the same PDE on the low-resolution grid, yielding $u_{\text{out}}^{\mathrm{lr}}$. Sparse observations are obtained by masking $u_{\text{out}}^{\mathrm{lr}}$ and adding Gaussian noise only at observed locations.

\begin{table*}[t]
\centering
\caption{Comparison of different models across PDE families. For 3D-Var and EnKF, we report the better result between the MAP and encoder initial priors, chosen by the lowest aggregate RMSE in each setting. MAE is reported for deterministic point estimates, and CRPS for probabilistic predictions. A lower value is more desirable for all metrics listed.}
\label{tab:methods_comparison_source_on}
\resizebox{\textwidth}{!}{
\begin{tabular}{l cccc cccc cccc}
\toprule
 & \multicolumn{4}{c}{Advection--Diffusion} & \multicolumn{4}{c}{Klein--Gordon} & \multicolumn{4}{c}{Helmholtz} \\
\cmidrule(lr){2-5} \cmidrule(lr){6-9} \cmidrule(lr){10-13}
Method & RMSE & PSD & MAE & CRPS & RMSE & PSD & MAE & CRPS & RMSE & PSD & MAE & CRPS \\
\midrule
LatentPDE-MAP (Ours) & 0.128 & 2.442 & -- & 0.087 & \textbf{0.601} & 3.952 & -- & \textbf{0.388} & 0.088 & 3.628 & -- & 0.071 \\
LatentPDE-Enc (Ours) & \textbf{0.103} & \textbf{1.570} & -- & \textbf{0.056} & 0.743 & 4.042 & -- & 0.458 & 0.052 & \textbf{3.433} & -- & \textbf{0.025} \\
\midrule \addlinespace
EnKF & 0.134 & 10.819 & -- & 0.104 & 0.626 & 3.736 & -- & 0.484 & 0.044 & 7.474 & -- & 0.031 \\
3D-Var & 0.112 & 7.322 & 0.104 & -- & 0.605 & \textbf{3.554} & 0.507 & -- & \textbf{0.038} & 4.265 & 0.028 & -- \\
PINN & 0.170 & 7.767 & 0.137 & -- & 0.732 & 6.113 & 0.573 & -- & 0.083 & 5.662 & 0.071 & -- \\
\bottomrule
\end{tabular}
}
\end{table*}

The benchmark comprises six regimes spanning three PDE families, whose functional forms can be found in Table \ref{tab:pde_equations}. Four regimes correspond to distinct parameter subregimes of advection--diffusion, while the remaining two instantiate the Klein--Gordon and Helmholtz families. Each regime specifies a PDE family, parameter ranges, and the standard deviation used to sample latent forcing coefficients. Full regime details are provided in Table~\ref{tab:regime_ranges}.

\begin{table}[h]
    \centering
    \caption{Summary of 2D homogeneous governing PDEs. Note that our experiments include an additional source term.}
    \label{tab:pde_equations}
    \begin{tabular}{@{}ll@{}}
        \toprule
        \textbf{Equation Name} & \textbf{Formulation} \\
        \midrule
        Advection--Diffusion    & $u_t + \mathbf{v} \cdot \nabla u = \kappa \Delta u$ \\
        Klein--Gordon           & $u_{tt} - \mathbf{c}^2 \nabla^2 u + m^2 u = 0$ \\
        Helmholtz              & $-\boldsymbol{\kappa}\nabla^2 u + k^2 u = u_0$ \\
        \bottomrule
    \end{tabular}
\end{table}

For each sample we construct a binary observation mask on the low-resolution grid with a prescribed sparsity $0<s\leq 1$. The training pipeline uses four mask types with sample-wise sparsity drawn uniformly from a range, while evaluation additionally uses structured OOD masks. Sample masks can be viewed in Appendix \ref{sec:appendix_maskgen}.

\subsection{Spectral Forcing Representation}
\label{subsec:forcing}

All three PDE families include an unknown spatially-varying, time-independent source term $f : \Omega \to \mathbb{R}$. Rather than learning or optimizing a forcing value at every grid point, we represent it by a truncated set of low-frequency Fourier coefficients.  Specifically, we retain the $K_f \times K_f$ lowest-wavenumber block of the two-dimensional real FFT, split into positive and negative row-frequency bands and stored as real and imaginary parts separately, yielding a forcing sub-vector $\mathbf{q}_f \in \mathbb{R}^{4K_f^2}$.  At decode time, $\mathbf{q}_f$ is zero-padded into an $H \times (W/2+1)$ half-spectrum and inverted via an inverse FFT to produce $f$ at the target resolution.

This design has two consequences.  First, the forcing is \emph{resolution-transferable} by construction, meaning that the same latent vector decodes to a physically consistent $f$ at any output grid size. This is because zero-padding a truncated spectrum is equivalent to ideal band-limited interpolation.  Second, limiting inference to $K_f$ modes acts as an implicit low-frequency prior on the source term, which prevents the latent from overfitting high-frequency observation noise. For a more mathematical explanation of the forcing term, we refer the reader to Appendix~\ref{sec:appendix_forcing}.

\subsection{Experimental Setup}
Unless otherwise noted, models are trained exclusively within their own PDE family and results are reported separately for each class. Each regime contributes $5000$ synthetic training examples. Training uses four different masks with $s \sim U[0.01,0.15]$, while evaluation fixes $s=5\%$. We evaluate $500$ test instances per regime. For LatentPDE, we draw $12$ posterior samples per test case (a discussion of how this number was derived can be found in Appendix~\ref{sec:appendix_ensemble_size}) and use the ensemble mean as the reconstruction.


 An important note in evaluating the performance of our framework lies in the choice between \textsc{LatentPDE-MAP} and \textsc{LatentPDE-Enc}. To account for the fact that in any realistic scenario we would not have access to the ground truth of the test set, we first compute the error on each PDE family using a separate validation set. Then, we select the variant with the lower validation error to use for the entirety of the held-out test set. Consequently, in any future reference to LatentPDE in our plots, it refers to the initialization mode that had the lower validation error. A full discussion of our setup can be found in Appendix~\ref{sec:appendix_train_test}.

\subsection{Baselines} \label{sec:baselines}
We consider four primary baselines: two standard data assimilation techniques, 3D-Var \cite{lorenc1986analysis} and the Ensemble Kalman Filter (EnKF) \cite{evensen1994sequential}, which analogously deal with incomplete data and reconstruction, the well-known broader framework of PINNs \cite{RAISSI2019686}, and a more recent state-of-the-art generative model FunDPS \cite{yao2025guided}. 

\begin{figure*}[h!]
    \centering
    \includegraphics[width=.9\textwidth]{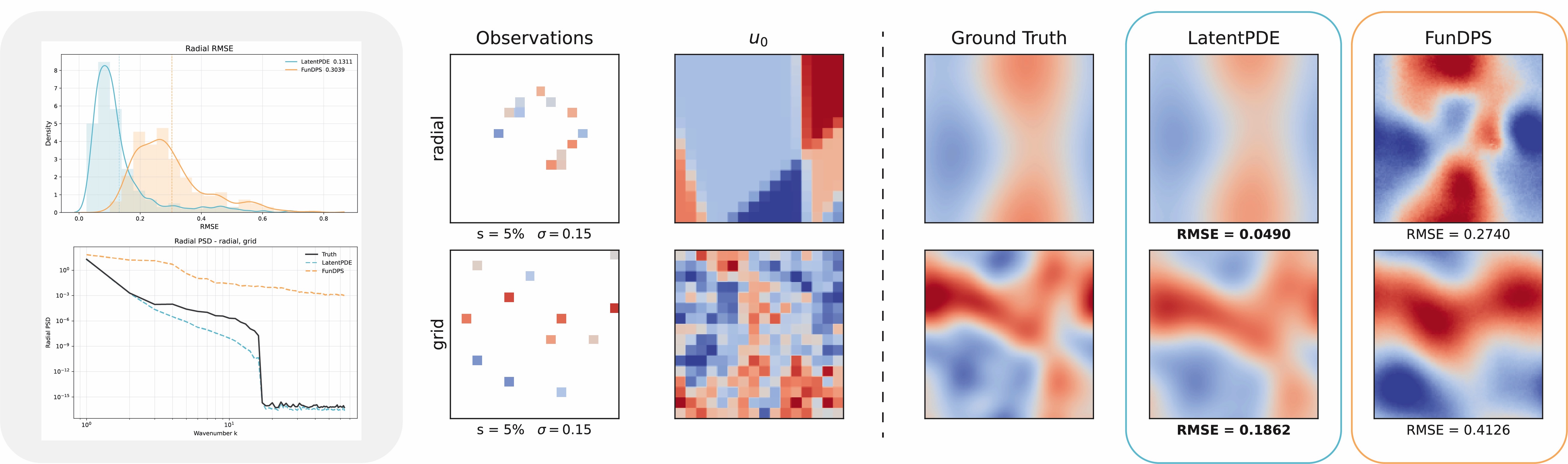}
    \caption{Comparison of LatentPDE and FunDPS reconstructions. FunDPS was trained on the same 20000 samples of advection--diffusion data as our model. Error distributions and spectral alignment curves over 500 samples are shown on the left. Top and bottom row show different individual reconstructions of radial and grid mask respectively.}
    \label{fig:fundps_comp}
\end{figure*}

To briefly explain the methodological differences between these baselines, 3D-Var combines prior knowledge usually referred to as the background with known observations  to obtain the optimal state or prediction of a field variable through a deterministic optimization problem. It relies on assumptions of Gaussian error distributions and isotropic background covariances \cite{barker2003three, weaver2001correlation}. As a probabilistic counterpart, EnKF utilizes a Monte Carlo approach, propagating an ensemble of model states forward to dynamically estimate background error covariances rather than relying on static assumptions \cite{katzfuss2016understanding}. Note that we test both MAP and encoder intializations as the background state for 3D-Var \& EnKF methods, then report the one yielding the lowest Root Mean Square Error (RMSE). PINNs use PDE residuals as regularizers within a composite loss function and softly constrain the network towards physically valid states. FunDPS uses a joint formulation of solution and coefficient spaces to solve forward and inverse problems using a functional space extension of diffusion models guided by PDE constraints.

For most of the baselines, there is no direct analogue to converting low-resolution inputs to high-resolution outputs as LatentPDE can. To accommodate this structural difference and ensure a competitive comparison, we supply the baseline methods with high-resolution inputs. While we recognize that FunDPS has upsampling capabilities, we also provide it with high-resolution data to remain consistent with other baselines. In contrast, LatentPDE operates strictly with low-resolution, sparse observations and the initial condition.  We report RMSE for pointwise accuracy and Power Spectral Density (PSD) log error to measure the recovery of multi-scale spatial frequencies \cite{sinclair2005empirical, harris2001multiscale}. For probabilistic posterior samplers, we report the Continuous Ranked Probability Score (CRPS) to assess uncertainty calibration against noise, and analogously the Mean Absolute Error (MAE) for deterministic models. Additional details on metrics can be found in Appendix \ref{sec:appendix_eval_metrics}.

\subsection{Sparse Reconstruction}
We present the reconstruction results of LatentPDE benchmarked against traditional methods like EnKF, 3D-Var, and PINNs. As shown in Figure \ref{fig:main_results}, LatentPDE is able to achieve the lowest RMSE across the board and qualitatively captures much of the finer-scale details present in the ground truth. While some of the baselines produce visually smooth fields, like in the PINN prediction for the advection--diffusion scenario, this over-regularization represents a failure to capture the true underlying physics. On the other hand, EnKF exhibits high-frequency artifacts, showing signs of aggressive over-fitting to the local noisy observation points.

A fuller collection of metrics can be found in Table \ref{tab:methods_comparison_source_on}. Overall, LatentPDE consistently surpasses other high-resolution baselines in matching the spectral characteristics of the ground truth fields, reflected in substantially lower PSD errors. Likewise, the framework reveals robust uncertainty quantification against injected noise as verified through the superior CRPS values. Because the DA methods share the same initializations as LatentPDE, these metrics provide further justification that the diffusion process plays a non-trivial role in refining the approximation. Instead of relying purely on a strong initial guess, LatentPDE is actively using the posterior to resolve fine-scale physical features that traditional DA methods cannot.

\begin{figure}[h!]
    \centering
    \includegraphics[width=.45\textwidth]{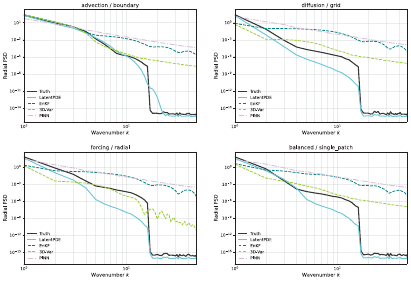}
    \caption{Radially averaged PSD across advection--diffusion subregimes and mask configurations. LatentPDE matches the ground truth energy even at the sharp high-frequency cutoff, whereas the baselines suffer from spectral bias.}
    \label{fig:psd_advdiff}
\end{figure}

To visually illustrate the spectral alignment with ground truth, Figure \ref{fig:psd_advdiff} plots the PSD curve for the four subregimes of the advection--diffusion equation. PSD intuitively measures how a field's variance is distributed across spatial scales ($k$), and therefore acts as a rigorous metric for structural realism. In analyzing the PSD of our reconstruction against that of the ground truth, we show that LatentPDE is successfully filtering out the injected noise and simultaneously recovering the lost sub-grid physical structures.

Moreover, we compare the reconstruction of LatentPDE with FunDPS for the inverse problem of the advection--diffusion equation on two distinct OOD masks with $s = 5\%$ and $\sigma=0.15$ in Figure \ref{fig:fundps_comp}. We demonstrate lower and tighter RMSE distribution compared to the noisy artifacts produced by FunDPS. The radial PSD on the bottom left further confirms this structural advantage. LatentPDE is much more aligned with the spectral decay of the ground truth, whereas FunDPS exhibits non-physical spectral energy across wavenumbers. A breakdown of runtimes for all models are provided in Appendix \ref{sec:appendix_compute}.

\subsection{Uncertainty Quantification}
\begin{figure}[h!]
    \centering
    \includegraphics[width=.45\textwidth]{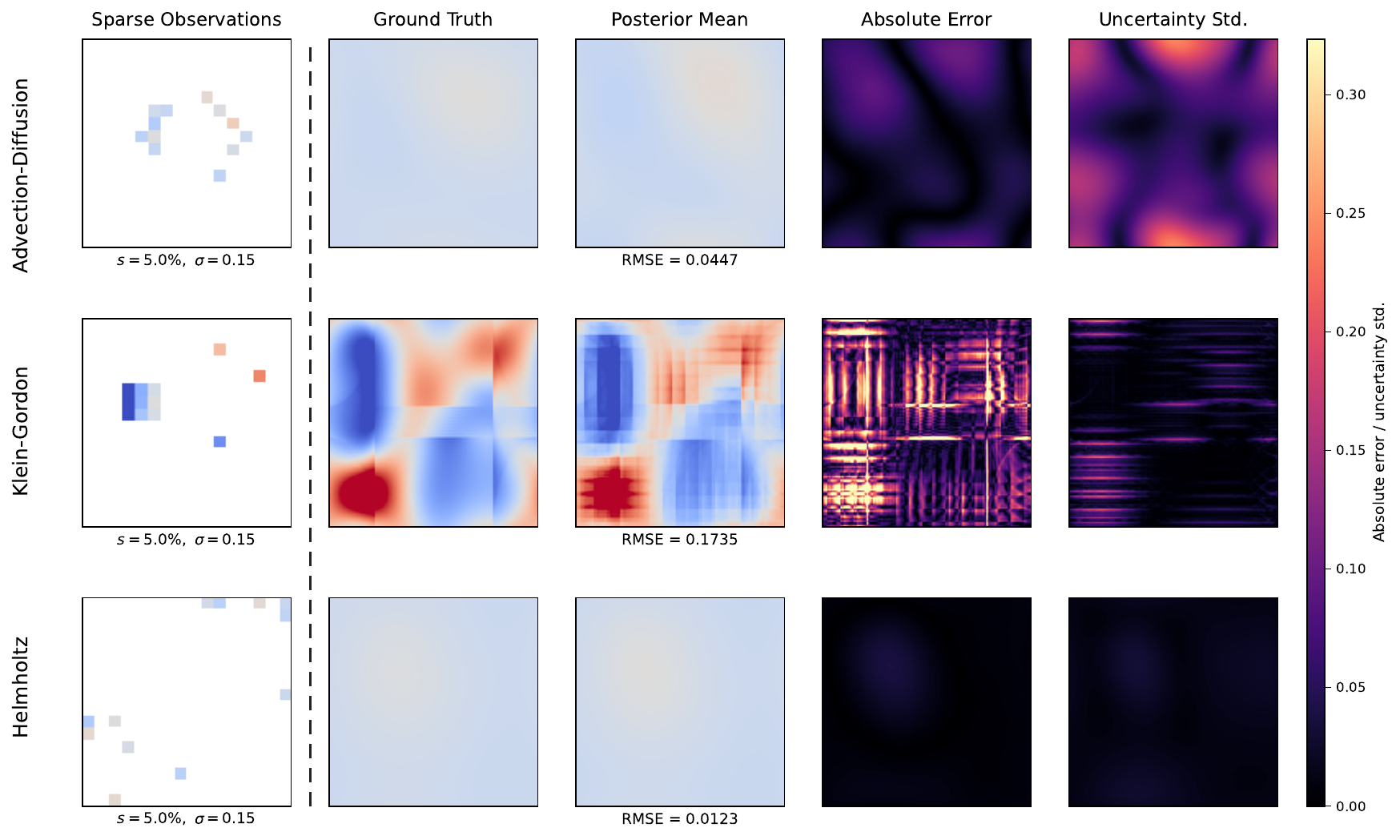}
    \caption{Ensemble mean and uncertainty quantification for LatentPDE across our three governing equations. The rightmost columns demonstrate that the generated uncertainty standard deviation maps spatially correlate to regions of elevated absolute error.}
    \label{fig:posterior_uncertainty}
\end{figure}

We can gain deeper insights into the robustness of the model by investigating the distribution of our generative outputs. As such, Figure \ref{fig:posterior_uncertainty} visualizes the ensemble mean reconstruction from LatentPDE along with uncertainty plots. Here, the standard deviation fields provide a spatial map of posterior spread, highlighting regions where the reconstruction is less certain. 


\begin{figure}[h!]
    \centering
    \includegraphics[width=.47\textwidth]{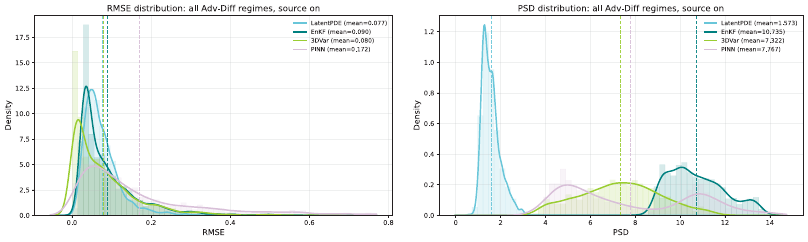}
    \caption{Metric distributions for RMSE and PSD for all subregimes of the advection--diffusion equation.}
    \label{fig:uq_advdiff}
\end{figure}

These high-uncertainty regions coincide qualitatively with larger absolute errors, indicating that the posterior ensemble captures meaningful localized failure modes under sparse conditioning. Beyond these spatial insights, encoding physical coefficients into our latents provides a restriction upon the solution space. As shown in Figure \ref{fig:uq_advdiff}, this results in tighter error bounds and better spectral fidelity than conventional methods when aggregated across multiple runs.

\section{Discussion}

A key takeaway from these evaluations is the distinct roles played by \textsc{LatentPDE-MAP} and \textsc{LatentPDE-Enc}. \textsc{LatentPDE-MAP} initializes directly against the masked residual of the test sample, and is therefore aligned with given observations but blind to the broader training distribution. On the contrary, \textsc{LatentPDE-Enc} initializes using an amortized prediction. This makes it so that regularities learned across the training set can be exploited to provide a better starting basis. Depending on the problem setup, either method may yield more favorable reconstructions. For example, when the inverse problem is severely underdetermined, the encoder's learned inductive bias allows the model to make predictions with assumptions inherited from the training data. Yet, when the observations are already informative enough, the encoder can introduce an unwanted bias, making MAP a better choice. In cases where the posterior is well captured by the data, both initializations provide valid guesses and yield similar results.

To summarize, neither strategy dominates unconditionally. Their relative performance reflects how much information the observations carry relative to the prior, and how well the training distribution covers the test geometry. 

\section{Conclusion}
In this work, we propose LatentPDE, a comprehensive framework for inverting the unknown coefficients of a governing PDE from sparse, noisy, and low-resolution observations. This is accomplished by enforcing physics within an interpretable latent space and a coupled spectral solver. We demonstrate improved structural realism via power spectral densities and reduced error. While LatentPDE yields promising results for linear PDEs, we intend to expand the framework to handle nonlinear dynamics with non-periodic boundary conditions and irregular domains. Another limitation is the inability to address \textit{equifinality}, wherein multiple distinct sets of coefficients can yield similarly plausible reconstructions for a given sparse input. Future works could condition upon multiple timesteps rather than a single snapshot to address this. Additionally, our reliance on a spectral solver as a decoder restricts the framework to smooth solutions. Thus, developing a more general decoder to accurately capture these abrupt dynamics is of interest to explore.

\section*{Impact Statement}
This work advances our ability to accurately model complex physical systems even when observational data is highly deficient in quality and/or quantity. Broadly, it can be used for the democratization of high-fidelity predictive modeling in under-resourced areas, especially in applications such as disaster management and extreme weather forecasting. However, as with any latent space framework, there remains an ethical need to maintain transparency about the underlying physical constraints and ensure that users understand the uncertainty bounds of generated results to prevent overconfident decision-making.




\bibliography{example_paper}

@article{kovachki2023neural,
  title={Neural operator: Learning maps between function spaces with applications to pdes},
  author={Kovachki, Nikola and Li, Zongyi and Liu, Burigede and Azizzadenesheli, Kamyar and Bhattacharya, Kaushik and Stuart, Andrew and Anandkumar, Anima},
  journal={Journal of Machine Learning Research},
  volume={24},
  number={89},
  pages={1--97},
  year={2023}
}

@article{huang2024operator,
  title={An operator learning perspective on parameter-to-observable maps},
  author={Huang, Daniel Zhengyu and Nelsen, Nicholas H and Trautner, Margaret},
  journal={arXiv preprint arXiv:2402.06031},
  year={2024}
}

@ARTICLE{baldan2023electromagnetic,
  author={Baldan, Marco and Di Barba, Paolo and Lowther, David A.},
  journal={IEEE Transactions on Magnetics}, 
  title={Physics-Informed Neural Networks for Inverse Electromagnetic Problems}, 
  year={2023},
  volume={59},
  number={5},
  pages={1-5},
  doi={10.1109/TMAG.2023.3247023}}

@article{zheng2024hompinns,
  title={HomPINNs: Homotopy physics-informed neural networks for solving the inverse problems of nonlinear differential equations with multiple solutions},
  author={Zheng, Haoyang and Huang, Yao and Huang, Ziyang and Hao, Wenrui and Lin, Guang},
  journal={Journal of computational physics},
  volume={500},
  pages={112751},
  year={2024},
  publisher={Elsevier}
}

@article{kashinath2021physics,
  title={Physics-informed machine learning: case studies for weather and climate modelling},
  author={Kashinath, Karthik and Mustafa, Mustafa and Albert, Adrian and Wu, Jean-Luc and Jiang, C and Esmaeilzadeh, Soheil and Azizzadenesheli, Kamyar and Wang, R and Chattopadhyay, Ashesh and Singh, Aakanksha and others},
  journal={Philosophical Transactions of the Royal Society A: Mathematical, Physical and Engineering Sciences},
  volume={379},
  number={2194},
  year={2021},
  publisher={The Royal Society}
}

@article{gao2021super,
  title={Super-resolution and denoising of fluid flow using physics-informed convolutional neural networks without high-resolution labels},
  author={Gao, Han and Sun, Luning and Wang, Jian-Xun},
  journal={Physics of Fluids},
  volume={33},
  number={7},
  year={2021},
  publisher={AIP Publishing}
}

@article{felippa2004introduction,
  title={Introduction to finite element methods},
  author={Felippa, Carlos A},
  journal={University of Colorado},
  volume={885},
  year={2004},
  publisher={Boulder CO}
}

@book{thomas2013numerical,
  title={Numerical partial differential equations: finite difference methods},
  author={Thomas, James William},
  volume={22},
  year={2013},
  publisher={Springer Science \& Business Media}
}

@book{shen2011spectral,
  title={Spectral methods: algorithms, analysis and applications},
  author={Shen, Jie and Tang, Tao and Wang, Li-Lian},
  volume={41},
  year={2011},
  publisher={Springer Science \& Business Media}
}

@article{lv2025hybrid,
  title={Hybrid wave height forecasting via integrated physics-based simulation and data-driven correction with contrastive feature fusion},
  author={Lv, Caichao and Song, Ning and Nie, Jie and Ye, Min and Liang, Xinyue and Jia, Dongning and Ni, Xin},
  journal={Applied Ocean Research},
  volume={162},
  pages={104729},
  year={2025},
  publisher={Elsevier}
}

@article{arnst2013hybrid,
  title={Hybrid sampling/spectral method for solving stochastic coupled problems},
  author={Arnst, Maarten and Soize, Christian and Ghanem, Roger},
  journal={SIAM/ASA Journal on Uncertainty Quantification},
  volume={1},
  number={1},
  pages={218--243},
  year={2013},
  publisher={SIAM}
}

@inproceedings{tikhonov1943stability,
  title={On the stability of inverse problems},
  author={Tikhonov, Andrey Nikolayevich and others},
  booktitle={Dokl. akad. nauk sssr},
  volume={39},
  number={5},
  pages={195--198},
  year={1943}
}

@article{Stuart_2010, title={Inverse problems: A Bayesian perspective}, volume={19}, DOI={10.1017/S0962492910000061}, journal={Acta Numerica}, author={Stuart, A. M.}, year={2010}, pages={451–559}}

@article{courtier4dvar,
author = {Courtier, Philippe and Talagrand, Olivier},
title = {Variational Assimilation of Meteorological Observations With the Adjoint Vorticity Equation. Ii: Numerical Results},
journal = {Quarterly Journal of the Royal Meteorological Society},
volume = {113},
number = {478},
pages = {1329-1347},
doi = {https://doi.org/10.1002/qj.49711347813},
url = {https://rmets.onlinelibrary.wiley.com/doi/abs/10.1002/qj.49711347813},
eprint = {https://rmets.onlinelibrary.wiley.com/doi/pdf/10.1002/qj.49711347813},
year = {1987}
}

@article{evensen1994sequential,
  title={Sequential data assimilation with a nonlinear quasi-geostrophic model using Monte Carlo methods to forecast error statistics},
  author={Evensen, Geir},
  journal={Journal of Geophysical Research: Oceans},
  volume={99},
  number={C5},
  pages={10143--10162},
  year={1994},
  publisher={Wiley Online Library}
}

@article{burgers1998,
  title   = {Analysis scheme in the ensemble {Kalman} filter},
  author  = {Burgers, Gerrit and van Leeuwen, Peter Jan and Evensen, Geir},
  journal = {Monthly Weather Review},
  volume  = {126},
  number  = {6},
  pages   = {1719--1724},
  year    = {1998}
}

@article{iglesias2013ensemble,
  title={Ensemble Kalman methods for inverse problems},
  author={Iglesias, Marco A and Law, Kody JH and Stuart, Andrew M},
  journal={Inverse Problems},
  volume={29},
  number={4},
  pages={045001},
  year={2013},
  publisher={IOP Publishing}
}

@article{RAISSI2019686,
title = {Physics-informed neural networks: A deep learning framework for solving forward and inverse problems involving nonlinear partial differential equations},
journal = {Journal of Computational Physics},
volume = {378},
pages = {686-707},
year = {2019},
issn = {0021-9991},
doi = {https://doi.org/10.1016/j.jcp.2018.10.045},
url = {https://www.sciencedirect.com/science/article/pii/S0021999118307125},
author = {M. Raissi and P. Perdikaris and G.E. Karniadakis}
}

@article{lu2019deeponet,
  title={Deeponet: Learning nonlinear operators for identifying differential equations based on the universal approximation theorem of operators},
  author={Lu, Lu and Jin, Pengzhan and Karniadakis, George Em},
  journal={arXiv preprint arXiv:1910.03193},
  year={2019}
}

@article{li2020fourier,
  title={Fourier neural operator for parametric partial differential equations},
  author={Li, Zongyi and Kovachki, Nikola and Azizzadenesheli, Kamyar and Liu, Burigede and Bhattacharya, Kaushik and Stuart, Andrew and Anandkumar, Anima},
  journal={arXiv preprint arXiv:2010.08895},
  year={2020}
}

@article{deistler2025simulation,
  title={Simulation-based inference: A practical guide},
  author={Deistler, Michael and Boelts, Jan and Steinbach, Peter and Moss, Guy and Moreau, Thomas and Gloeckler, Manuel and Rodrigues, Pedro LC and Linhart, Julia and Lappalainen, Janne K and Miller, Benjamin Kurt and others},
  journal={arXiv preprint arXiv:2508.12939},
  year={2025}
}

@article{ward2022robust,
  title={Robust neural posterior estimation and statistical model criticism},
  author={Ward, Daniel and Cannon, Patrick and Beaumont, Mark and Fasiolo, Matteo and Schmon, Sebastian},
  journal={Advances in Neural Information Processing Systems},
  volume={35},
  pages={33845--33859},
  year={2022}
}

@article{kingma2013auto,
  title={Auto-encoding variational bayes},
  author={Kingma, Diederik P and Welling, Max},
  journal={arXiv preprint arXiv:1312.6114},
  year={2013}
}

@inproceedings{sohl2015deep,
  title={Deep unsupervised learning using nonequilibrium thermodynamics},
  author={Sohl-Dickstein, Jascha and Weiss, Eric and Maheswaranathan, Niru and Ganguli, Surya},
  booktitle={International conference on machine learning},
  pages={2256--2265},
  year={2015},
  organization={pmlr}
}

@article{ho2020denoising,
  title={Denoising diffusion probabilistic models},
  author={Ho, Jonathan and Jain, Ajay and Abbeel, Pieter},
  journal={Advances in neural information processing systems},
  volume={33},
  pages={6840--6851},
  year={2020}
}

@inproceedings{kim2018disentangling,
  title={Disentangling by factorising},
  author={Kim, Hyunjik and Mnih, Andriy},
  booktitle={International conference on machine learning},
  pages={2649--2658},
  year={2018},
  organization={PMLR}
}

@article{hinz2018inferencing,
  title={Inferencing based on unsupervised learning of disentangled representations},
  author={Hinz, Tobias and Wermter, Stefan},
  journal={arXiv preprint arXiv:1803.02627},
  year={2018}
}

@article{wang2024disentangled,
  title={Disentangled representation learning},
  author={Wang, Xin and Chen, Hong and Tang, Si'ao and Wu, Zihao and Zhu, Wenwu},
  journal={IEEE Transactions on Pattern Analysis and Machine Intelligence},
  volume={46},
  number={12},
  pages={9677--9696},
  year={2024},
  publisher={IEEE}
}

@inproceedings{rombach2022high,
  title={High-resolution image synthesis with latent diffusion models},
  author={Rombach, Robin and Blattmann, Andreas and Lorenz, Dominik and Esser, Patrick and Ommer, Bj{\"o}rn},
  booktitle={Proceedings of the IEEE/CVF conference on computer vision and pattern recognition},
  pages={10684--10695},
  year={2022}
}

@article{bastek2024physics,
  title={Physics-informed diffusion models},
  author={Bastek, Jan-Hendrik and Sun, WaiChing and Kochmann, Dennis M},
  journal={arXiv preprint arXiv:2403.14404},
  year={2024}
}

@article{chefer2023hidden,
  title={The hidden language of diffusion models},
  author={Chefer, Hila and Lang, Oran and Geva, Mor and Polosukhin, Volodymyr and Shocher, Assaf and Irani, Michal and Mosseri, Inbar and Wolf, Lior},
  journal={arXiv preprint arXiv:2306.00966},
  year={2023}
}

@inproceedings{perez2018film,
  title={Film: Visual reasoning with a general conditioning layer},
  author={Perez, Ethan and Strub, Florian and De Vries, Harm and Dumoulin, Vincent and Courville, Aaron},
  booktitle={Proceedings of the AAAI conference on artificial intelligence},
  volume={32},
  number={1},
  year={2018}
}

@article{lorenc1986analysis,
author = {Lorenc, A. C.},
title = {Analysis methods for numerical weather prediction},
journal = {Quarterly Journal of the Royal Meteorological Society},
volume = {112},
number = {474},
pages = {1177-1194},
doi = {https://doi.org/10.1002/qj.49711247414},
url = {https://rmets.onlinelibrary.wiley.com/doi/abs/10.1002/qj.49711247414},
eprint = {https://rmets.onlinelibrary.wiley.com/doi/pdf/10.1002/qj.49711247414},
year = {1986}
}

@article{barker2003three,
  title={A three-dimensional variational (3DVAR) data assimilation system for use with MM5},
  author={Barker, DM and Huang, Wei and Guo, Yong-Run and Bourgeois, Al},
  journal={NCAR Tech Note},
  volume={68},
  year={2003},
  publisher={NCAR/TN-453s1R}
}

@article{weaver2001correlation,
  title={Correlation modelling on the sphere using a generalized diffusion equation},
  author={Weaver, Anthony and Courtier, Philippe},
  journal={Quarterly Journal of the Royal Meteorological Society},
  volume={127},
  number={575},
  pages={1815--1846},
  year={2001},
  publisher={Wiley Online Library}
}

@article{katzfuss2016understanding,
  title={Understanding the ensemble Kalman filter},
  author={Katzfuss, Matthias and Stroud, Jonathan R and Wikle, Christopher K},
  journal={The American Statistician},
  volume={70},
  number={4},
  pages={350--357},
  year={2016},
  publisher={Taylor \& Francis}
}

@article{sinclair2005empirical,
  title={Empirical Mode Decomposition in 2-D space and time: a tool for space-time rainfall analysis and nowcasting},
  author={Sinclair, S and Pegram, GGS},
  journal={Hydrology and Earth System Sciences},
  volume={9},
  number={3},
  pages={127--137},
  year={2005},
  publisher={Copernicus Publications G{\"o}ttingen, Germany}
}

@article {harris2001multiscale,
      author = "Daniel  Harris and Efi  Foufoula-Georgiou and Kelvin K.  Droegemeier and Jason J.  Levit",
      title = "Multiscale Statistical Properties of a High-Resolution Precipitation Forecast",
      journal = "Journal of Hydrometeorology",
      year = "2001",
      publisher = "American Meteorological Society",
      address = "Boston MA, USA",
      volume = "2",
      number = "4",
      doi = "10.1175/1525-7541(2001)002<0406:MSPOAH>2.0.CO;2",
      pages=      "406 - 418",
      url = "https://journals.ametsoc.org/view/journals/hydr/2/4/1525-7541_2001_002_0406_mspoah_2_0_co_2.xml"
}

@article{yao2025guided,
  title={Guided diffusion sampling on function spaces with applications to pdes},
  author={Yao, Jiachen and Mammadov, Abbas and Berner, Julius and Kerrigan, Gavin and Ye, Jong Chul and Azizzadenesheli, Kamyar and Anandkumar, Anima},
  journal={arXiv preprint arXiv:2505.17004},
  year={2025}
}

@inproceedings{ho2022classifierfree,
  title     = {Classifier-Free Diffusion Guidance},
  author    = {Ho, Jonathan and Salimans, Tim},
  booktitle = {NeurIPS Workshop on Deep Generative Models and Downstream Applications},
  year      = {2021}
}

@inproceedings{chung2022dps,
  title     = {Diffusion Posterior Sampling for General Noisy Inverse Problems},
  author    = {Chung, Hyungjin and Kim, Jeongsol and Mccann, Michael T and Klasky, Marc L and Ye, Jong Chul},
  booktitle = {International Conference on Learning Representations},
  year      = {2023}
}
\bibliographystyle{icml2026}

\newpage
\appendix
\onecolumn
\section*{Appendix}

\section{Latent Parameterization Details}
\label{sec:appendix_latent}

The first three raw latent coordinates $(r_1, r_2, r_3)$ are mapped to family-specific physical parameters via interval-constrained sigmoids:
\begin{equation}
    \theta_j = a_{\mathfrak{f},j} + (b_{\mathfrak{f},j} - a_{\mathfrak{f},j})\,\sigma(r_j), \quad j = 1,2,3,
\end{equation}
where $\sigma(\cdot)$ is the logistic sigmoid and $[(a_{\mathfrak{f},j}, b_{\mathfrak{f},j})]_{j=1}^3$ are family-specific parameter bounds. The three coordinates decode to $(v_x, v_y, \kappa)$ for advection--diffusion, $(c_x, c_y, m)$ for Klein--Gordon, and $(\kappa_x, \kappa_y, k)$ for Helmholtz.

During inference, raw latents are normalized coordinate-wise as
\begin{equation}
    \mathbf{z} = \frac{\mathbf{z}_\mathrm{raw} - \boldsymbol{\mu}}{\boldsymbol{\sigma}},
\end{equation}
where $(\boldsymbol{\mu}, \boldsymbol{\sigma})$ are branch-specific statistics computed over the training set.

\section{Additional Details on Problem Setup}
\label{sec:appendix_problem_setup}

All implemented PDE families fit the general operator form
\begin{equation}\label{eq:general_operator}
    \mathcal{P}_{\mathfrak{f},\boldsymbol{\theta}}(D_t, \nabla)\, u = \mathcal{B}_{\mathfrak{f},\boldsymbol{\theta}}[u_0] + f,
\end{equation}
where $\mathcal{P}_{\mathfrak{f},\boldsymbol{\theta}}$ is a linear constant-coefficient space-time differential operator and $\mathcal{B}_{\mathfrak{f},\boldsymbol{\theta}}$ encodes initial-condition dependence. For evolutionary families this is posed on $\Omega \times [0,T]$; for Helmholtz it is a static boundary-value problem on $\Omega$.

Because every family is linear, constant-coefficient, and periodic, spatial Fourier modes decouple and each evolves via scalar transfer multipliers $G_\mathfrak{f}(\mathbf{k}, T; \boldsymbol{\theta})$ and $H_\mathfrak{f}(\mathbf{k}, T; \boldsymbol{\theta})$ as in Eq.~\eqref{eq:spectral_transfer}. The family-specific forms are as follows.

\paragraph{Advection--Diffusion.} With velocity $(v_x, v_y)$ and diffusivity $\kappa$, i.e.\ $\boldsymbol{\theta} = (v_x, v_y, \kappa)$:
\begin{align}
    G_\mathfrak{f}(\mathbf{k}, T; \boldsymbol{\theta}) &= \exp\!\left(-\left(2\pi i (k_x v_x + k_y v_y) + 4\pi^2 \kappa |\mathbf{k}|^2\right) T\right), \\
    H_\mathfrak{f}(\mathbf{k}, T; \boldsymbol{\theta}) &= \frac{G_\mathfrak{f}(\mathbf{k},T;\boldsymbol{\theta}) - 1}{-\left(2\pi i(k_x v_x + k_y v_y) + 4\pi^2\kappa|\mathbf{k}|^2\right)},
\end{align}
with $H_\mathfrak{f} = T$ when the denominator vanishes.

\paragraph{Klein--Gordon.} With wave speeds $(c_x, c_y)$ and mass $m$, i.e.\ $\boldsymbol{\theta} = (c_x, c_y, m)$, defining $\omega^2 = 4\pi^2(c_x^2 k_x^2 + c_y^2 k_y^2) + m^2$:
\begin{align}
    G_\mathfrak{f}(\mathbf{k}, T; \boldsymbol{\theta}) &= \cos(\omega T), \\
    H_\mathfrak{f}(\mathbf{k}, T; \boldsymbol{\theta}) &= \frac{\sin(\omega T)}{\omega}.
\end{align}

\paragraph{Helmholtz.} With anisotropic diffusivities $(\kappa_x, \kappa_y)$ and wavenumber $k$, i.e.\ $\boldsymbol{\theta} = (\kappa_x, \kappa_y, k)$:
\begin{align}
    G_\mathfrak{f}(\mathbf{k}; \boldsymbol{\theta}) &= 0, \\
    H_\mathfrak{f}(\mathbf{k}; \boldsymbol{\theta}) &= \frac{-1}{4\pi^2(\kappa_x k_x^2 + \kappa_y k_y^2) - k^2},
\end{align}
so the output depends only on the forcing (no time dependence). A full derivation of each of these can be found in Section~\ref{sec:appendix_pde_derivations}.

\section{PDE Families and Spectral Derivations}
\label{sec:appendix_pde_derivations}

This section derives the family-agnostic operator formulation to the three PDE families that we study and formalizes the exact spectral transfer functions used by the decoder.

\subsection{General Fourier Reduction of Linear Periodic Operator Families}
\label{subsec:appendix_general_fourier}

Let $\mathbf{x}=(x,y)\in\mathbb{T}^{2}$. Any sufficiently regular periodic field
admits the Fourier expansion
\begin{equation}
    u(\mathbf{x},t)
    =
    \sum_{\mathbf{n}\in\mathbb{Z}^{2}}
    \widehat{u}_{\mathbf{n}}(t)\,
    e^{2\pi i \mathbf{n}\cdot\mathbf{x}}.
\end{equation}
Equivalently, if we define the physical wavevector
$\mathbf{k}=2\pi\mathbf{n}=(k_{x},k_{y})$, then the spatial derivatives act on
each Fourier mode by
\begin{equation}
    \widehat{\partial_{x}u}(\mathbf{k},t) = i k_{x}\widehat{u}(\mathbf{k},t),
    \qquad
    \widehat{\partial_{y}u}(\mathbf{k},t) = i k_{y}\widehat{u}(\mathbf{k},t),
\end{equation}
and more generally
\begin{equation}
    \widehat{\partial_{x}^{\alpha_{1}}\partial_{y}^{\alpha_{2}}u}
    (\mathbf{k},t)
    =
    (ik_{x})^{\alpha_{1}}(ik_{y})^{\alpha_{2}}
    \widehat{u}(\mathbf{k},t).
\end{equation}

Suppose
\begin{equation}
    \mathcal{P}_{\mathfrak{f},\boldsymbol{\theta}}(D_{t},\nabla)
    =
    \sum_{m=0}^{m_{\max}}
    \sum_{|\alpha|\le r}
    c_{m,\alpha}^{(\mathfrak{f})}(\boldsymbol{\theta})\,
    \partial_{t}^{m}\partial_{x}^{\alpha_{1}}\partial_{y}^{\alpha_{2}},
    \qquad
    \alpha=(\alpha_{1},\alpha_{2}),
\end{equation}
with coefficients independent of $(\mathbf{x},t)$. Applying the spatial Fourier transform to Eq.~\eqref{eq:general_operator} yields, mode by mode,
\begin{equation}
    \sum_{m=0}^{m_{\max}}
    p_{m}^{(\mathfrak{f})}(\mathbf{k};\boldsymbol{\theta})\,
    \frac{d^{m}}{dt^{m}}
    \widehat{u}(\mathbf{k},t)
    =
    \widehat{b}_{\mathfrak{f}}(\mathbf{k},t;u_{0},\boldsymbol{\theta})
    +
    \widehat{f}(\mathbf{k}),
    \label{eq:appendix_modewise_reduction}
\end{equation}
where
\begin{equation}
    p_{m}^{(\mathfrak{f})}(\mathbf{k};\boldsymbol{\theta})
    =
    \sum_{|\alpha|\le r}
    c_{m,\alpha}^{(\mathfrak{f})}(\boldsymbol{\theta})\,
    (ik_{x})^{\alpha_{1}}(ik_{y})^{\alpha_{2}}.
\end{equation}
Thus every spatial mode decouples. In the study, the resulting modewise problem is either (1) a first-order scalar ODE (advection--diffusion), (2) a second-order scalar ODE (Klein--Gordon), or (3) an algebraic equation (Helmholtz family). This decoupling is the reason we are able to use a single FFT-based solve interface without changing the rest of the inference stack.

\subsection{Advection--Diffusion Family}
\label{subsec:appendix_advection_diffusion}

The advection--diffusion family is
\begin{equation}
    \partial_{t}u(\mathbf{x},t)
    +
    \mathbf{v}\cdot\nabla u(\mathbf{x},t)
    =
    \kappa \Delta u(\mathbf{x},t)
    +
    f(\mathbf{x}),
    \qquad
    \mathbf{v}=(v_{x},v_{y}),
    \qquad
    \kappa > 0,
    \label{eq:appendix_advection_diffusion_pde}
\end{equation}
with coefficient vector
\(
\boldsymbol{\theta}_{\mathsf{AD}}=(v_{x},v_{y},\kappa)
\),
with initial condition $u(\mathbf{x},0)=u_{0}(\mathbf{x})$ and periodic
boundary conditions. Fourier transforming in space gives
\begin{equation}
    \partial_{t}\widehat{u}(\mathbf{k},t)
    +
    i(v_{x}k_{x} + v_{y}k_{y})\widehat{u}(\mathbf{k},t)
    =
    -\kappa(k_{x}^{2}+k_{y}^{2})\widehat{u}(\mathbf{k},t)
    +
    \widehat{f}(\mathbf{k}),
\end{equation}
or equivalently
\begin{equation}
    \partial_{t}\widehat{u}(\mathbf{k},t)
    =
    \lambda_{\mathsf{AD}}(\mathbf{k})\,
    \widehat{u}(\mathbf{k},t)
    +
    \widehat{f}(\mathbf{k}),
    \label{eq:appendix_ad_mode_ode}
\end{equation}
with mode symbol
\begin{equation}
    \lambda_{\mathsf{AD}}(\mathbf{k})
    =
    -i(v_{x}k_{x}+v_{y}k_{y})
    -
    \kappa(k_{x}^{2}+k_{y}^{2}).
\end{equation}

Equation~\eqref{eq:appendix_ad_mode_ode} is a scalar affine ODE. Using the
integrating factor $e^{-\lambda_{\mathsf{AD}}(\mathbf{k})t}$,
\begin{equation}
    \frac{d}{dt}
    \left[
    e^{-\lambda_{\mathsf{AD}}(\mathbf{k})t}
    \widehat{u}(\mathbf{k},t)
    \right]
    =
    e^{-\lambda_{\mathsf{AD}}(\mathbf{k})t}\widehat{f}(\mathbf{k}).
\end{equation}
Integrating from $0$ to $T$ and using
$\widehat{u}(\mathbf{k},0)=\widehat{u}_{0}(\mathbf{k})$ gives
\begin{equation}
    \widehat{u}(\mathbf{k},T)
    =
    e^{\lambda_{\mathsf{AD}}(\mathbf{k})T}\,
    \widehat{u}_{0}(\mathbf{k})
    +
    \left(
    \int_{0}^{T}
    e^{\lambda_{\mathsf{AD}}(\mathbf{k})(T-s)}\,ds
    \right)
    \widehat{f}(\mathbf{k}).
\end{equation}
Evaluating the integral yields
\begin{equation}
    \widehat{u}(\mathbf{k},T)
    =
    e^{\lambda_{\mathsf{AD}}(\mathbf{k})T}\,
    \widehat{u}_{0}(\mathbf{k})
    +
    \frac{
    e^{\lambda_{\mathsf{AD}}(\mathbf{k})T} - 1
    }{
    \lambda_{\mathsf{AD}}(\mathbf{k})
    }\,
    \widehat{f}(\mathbf{k}).
    \label{eq:appendix_ad_transfer}
\end{equation}
Hence the family-specific spectral transfer functions are
\begin{equation}
    G_{\mathsf{AD}}(\mathbf{k},T;\boldsymbol{\theta}_{\mathsf{AD}})
    =
    e^{\lambda_{\mathsf{AD}}(\mathbf{k})T},
    \qquad
    H_{\mathsf{AD}}(\mathbf{k},T;\boldsymbol{\theta}_{\mathsf{AD}})
    =
    \frac{e^{\lambda_{\mathsf{AD}}(\mathbf{k})T}-1}
    {\lambda_{\mathsf{AD}}(\mathbf{k})}.
\end{equation}

At $\lambda_{\mathsf{AD}}(\mathbf{k})=0$, the forcing transfer function is defined by continuity:
\begin{equation}
    \lim_{\lambda\to 0}\frac{e^{\lambda T}-1}{\lambda} = T.
\end{equation}
The implementation applies this limiting value near zero to avoid numerical instability at very small denominators. We note that this is an exact continuous limit of the mode-wise transfer function and therefore preserves the mathematical definition of the decoder.

\subsection{Klein--Gordon Family}
\label{subsec:appendix_klein_gordon}

The Klein--Gordon family is given by
\begin{equation}
    \partial_{tt}u(\mathbf{x},t) - c_{x}^{2}\partial_{xx}u(\mathbf{x},t) - c_{y}^{2}\partial_{yy}u(\mathbf{x},t) + m^{2}u(\mathbf{x},t) = f(\mathbf{x}),
    \label{eq:appendix_kg_pde}
\end{equation}
with coefficient vector \(\boldsymbol{\theta}_{\mathsf{KG}} (c_{x},c_{y},m)\), and initial displacement and velocity
\begin{equation}
    u(\mathbf{x},0)=u_{0}(\mathbf{x}),\quad     \partial_{t}u(\mathbf{x},0)=0.
\end{equation}
The zero-velocity choice is a modeling decision induced by the current latent state, which stores only one initial field rather than an initial displacement-velocity pair.

Fourier transforming Eq.~\eqref{eq:appendix_kg_pde} yields
\begin{equation}
    \partial_{tt}\widehat{u}(\mathbf{k},t) + \omega_{\mathsf{KG}}(\mathbf{k})^{2} \widehat{u}(\mathbf{k},t) = \widehat{f}(\mathbf{k}),
    \label{eq:appendix_kg_mode_ode}
\end{equation}
where
\begin{equation}
    \omega_{\mathsf{KG}}(\mathbf{k})^{2} = c_{x}^{2}k_{x}^{2} + c_{y}^{2}k_{y}^{2} + m^{2}.
\end{equation}
Equation~\eqref{eq:appendix_kg_mode_ode} is a forced harmonic oscillator for each spatial mode. The homogeneous solution is then given by
\begin{equation}
    \widehat{u}_{h}(\mathbf{k},t)
    =
    A(\mathbf{k})\cos\bigl(\omega_{\mathsf{KG}}(\mathbf{k})t\bigr)
    +
    B(\mathbf{k})\sin\bigl(\omega_{\mathsf{KG}}(\mathbf{k})t\bigr).
\end{equation}
Because the forcing is time-independent, a particular solution is
\begin{equation}
    \widehat{u}_{p}(\mathbf{k},t) = \frac{\widehat{f}(\mathbf{k})} {\omega_{\mathsf{KG}}(\mathbf{k})^{2}}.
\end{equation}
Hence
\begin{equation}
    \widehat{u}(\mathbf{k},t) = A(\mathbf{k})\cos\bigl(\omega_{\mathsf{KG}}(\mathbf{k})t\bigr) + B(\mathbf{k})\sin\bigl(\omega_{\mathsf{KG}}(\mathbf{k})t\bigr) + \frac{\widehat{f}(\mathbf{k})} {\omega_{\mathsf{KG}}(\mathbf{k})^{2}}.
\end{equation}
Imposing the initial conditions gives
\begin{equation}
    A(\mathbf{k}) = \widehat{u}_{0}(\mathbf{k}) - \frac{\widehat{f}(\mathbf{k})}{\omega_{\mathsf{KG}}(\mathbf{k})^{2}}, \qquad B(\mathbf{k})=0,
\end{equation}
and therefore
\begin{equation}
    \widehat{u}(\mathbf{k},T) = \cos\bigl(\omega_{\mathsf{KG}}(\mathbf{k})T\bigr) \widehat{u}_{0}(\mathbf{k}) + \frac{1- \cos\bigl(\omega_{\mathsf{KG}}(\mathbf{k})T\bigr)}{\omega_{\mathsf{KG}}(\mathbf{k})^{2}}\widehat{f}(\mathbf{k}).
    \label{eq:appendix_kg_transfer}
\end{equation}
Thus
\begin{equation}
    G_{\mathsf{KG}}(\mathbf{k},T;\boldsymbol{\theta}_{\mathsf{KG}}) = \cos\bigl(\omega_{\mathsf{KG}}(\mathbf{k})T\bigr),
    \qquad H_{\mathsf{KG}}(\mathbf{k},T;\boldsymbol{\theta}_{\mathsf{KG}}) = \frac{1 - \cos\bigl(\omega_{\mathsf{KG}}(\mathbf{k})T\bigr) }{\omega_{\mathsf{KG}}(\mathbf{k})^{2}}.
\end{equation}

When $\omega_{\mathsf{KG}}(\mathbf{k})$ is very small, the forcing transfer function is again replaced by its limiting value:
\begin{equation}
    \lim_{\omega\to 0} \frac{1-\cos(\omega T)}{\omega^{2}} = \frac{T^{2}}{2}.
\end{equation}
 We note that this limit again preserves the mathematical definition of the decoder. Although the default parameter ranges keep $m$ bounded away from zero, our implementation retains this limit for numerical robustness.
 
\subsection{Helmholtz Family}
\label{subsec:appendix_helmholtz}

The Helmholtz family is a static equation of the form
\begin{equation}
    -\kappa_{x}\partial_{xx}u(\mathbf{x})
    -
    \kappa_{y}\partial_{yy}u(\mathbf{x})
    +
    k^{2} u(\mathbf{x})
    =
    u_{0}(\mathbf{x}) + f(\mathbf{x}),
    \label{eq:appendix_helmholtz_pde}
\end{equation}
with coefficient vector
\(
\boldsymbol{\theta}_{\mathsf{H}}=(\kappa_{x},\kappa_{y},k)
\),
with periodic boundary conditions. This is an anisotropic forced Helmholtz form in which the supplied initial condition appears on the right-hand side as a static source term.

Applying the spatial Fourier transform gives the algebraic relation
\begin{equation}
    \left[
    \kappa_{x}k_{x}^{2}
    +
    \kappa_{y}k_{y}^{2}
    +
    k^{2}
    \right]
    \widehat{u}(\mathbf{k})
    =
    \widehat{u}_{0}(\mathbf{k})
    +
    \widehat{f}(\mathbf{k}).
\end{equation}
Therefore
\begin{equation}
    \widehat{u}(\mathbf{k})
    =
    \frac{
    \widehat{u}_{0}(\mathbf{k}) + \widehat{f}(\mathbf{k})
    }{
    \kappa_{x}k_{x}^{2} + \kappa_{y}k_{y}^{2} + k^{2}
    }.
    \label{eq:appendix_helmholtz_transfer}
\end{equation}
In the notation of Eq.~\eqref{eq:spectral_transfer},
\begin{equation}
    G_{\mathsf{H}}(\mathbf{k};\boldsymbol{\theta}_{\mathsf{H}})
    =
    \frac{1}{
    \kappa_{x}k_{x}^{2} + \kappa_{y}k_{y}^{2} + k^{2}
    },
    \qquad
    H_{\mathsf{H}}(\mathbf{k};\boldsymbol{\theta}_{\mathsf{H}})
    =
    \frac{1}{
    \kappa_{x}k_{x}^{2} + \kappa_{y}k_{y}^{2} + k^{2}
    }.
\end{equation}
Because $\kappa_{x},\kappa_{y},k > 0$, the denominator is strictly positive and the solve is well posed at every Fourier mode, including the zero mode.

\section{Architecture Details}
\label{sec:appendix_architecture}

\paragraph{Sparse observation encoder.}
A four-channel input $[y, M, d(M), \rho(M)]$ is processed by a six-layer CNN (channels $4\to64\to128\to256$, two stride-2 convolutions, GroupNorm + GELU throughout) followed by global average pooling and a two-layer MLP, yielding $\mathbf{c}_{\mathrm{obs}} \in \mathbb{R}^{256}$.  This module is instantiated separately but with the same architecture in both the inference encoder and the denoiser.

\paragraph{Inference encoder.}
The encoder concatenates $[\mathbf{c}_{\mathrm{obs}}, \mathbf{c}_{u_0}]$ and passes through a four-layer MLP (hidden dim $512$, LayerNorm + GELU) projecting to $\mathbb{R}^{d_z}$. This is trained for 400 epochs.

\paragraph{Diffusion denoiser.}
The noisy latent $\mathbf{z}_t$ is projected to hidden dim $512$ and passed through six residual blocks (Linear + LayerNorm + GELU), each conditioned via a FiLM layer~\cite{perez2018film} on $\mathbf{c} = [\mathbf{c}_{\mathrm{obs}},\, \mathbf{c}_{u_0},\, \mathbf{c}_t] \in \mathbb{R}^{512}$, where $\mathbf{c}_t$ is a sinusoidal timestep embedding projected through a two-layer MLP.  A final linear layer maps back to $\mathbb{R}^{d_z}$. This is trained for 600 epochs.

\section{Synthetic Data Generation}

Synthetic data are generated separately for each regime, where a regime specifies a PDE family, three parameter ranges, and a forcing scale. Let $u_0^{\mathrm{hr}} \in \mathbb{R}^{H \times W}$ denote the high-resolution initial condition on the unit square.

\begin{figure}[h!]
    \centering
    \includegraphics[width=0.6\linewidth]{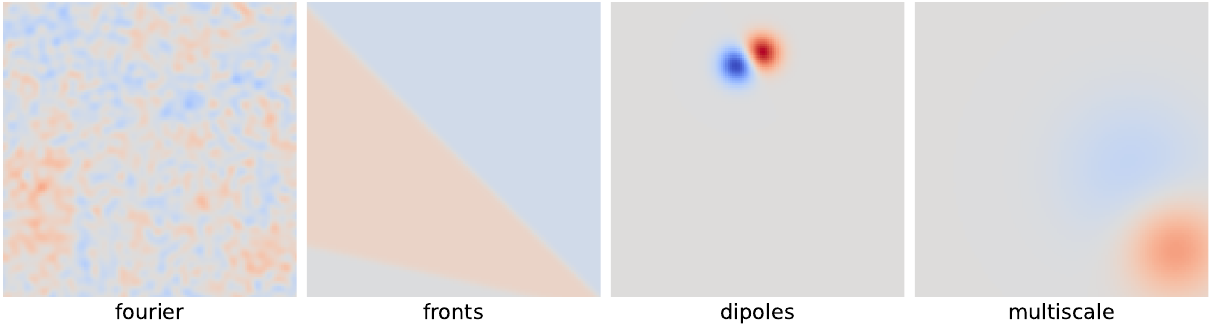}
    \caption{Representative initial conditions used in our study. These were chosen to provide challenging initial states with diverse spatial regularity and frequency content.}
    \label{fig:ic_viz}
\end{figure}

\paragraph{Initial conditions.}
Initial conditions are generated in the following four ways: broadband random Fourier field, sharp transition fronts, localized vortex dipoles, and multiscale Gaussian-Fourier fields. All initial conditions are standardized to zero mean and unit variance. Sample visualizations of each of these can be found in Figure~\ref{fig:ic_viz}.

\paragraph{Parameter and forcing sampling.}
Given regime $r$, each of the three physical coefficients is sampled independently from a regime-specific interval $[\ell_j^{(r)}, u_j^{(r)}]$. We first draw $\xi_j \sim U(0,1)$ and store a raw latent coordinate
\begin{equation}
z_j = \log \frac{\xi_j}{1-\xi_j},
\end{equation}
then decode
\begin{equation}
\theta_j
=
\ell_j^{(r)}
+
\bigl(u_j^{(r)} - \ell_j^{(r)}\bigr)\sigma(z_j),
\end{equation}
so that $\theta_j$ is uniform on $[\ell_j^{(r)}, u_j^{(r)}]$. When forcing is enabled, the latent forcing coefficients are sampled i.i.d.\ Gaussian with regime-specific standard deviation $\tau_r$. 

\subsection{Mask Generation} \label{sec:appendix_maskgen}
\begin{figure*}[h!]
    \centering
    \includegraphics[width=.6\textwidth]{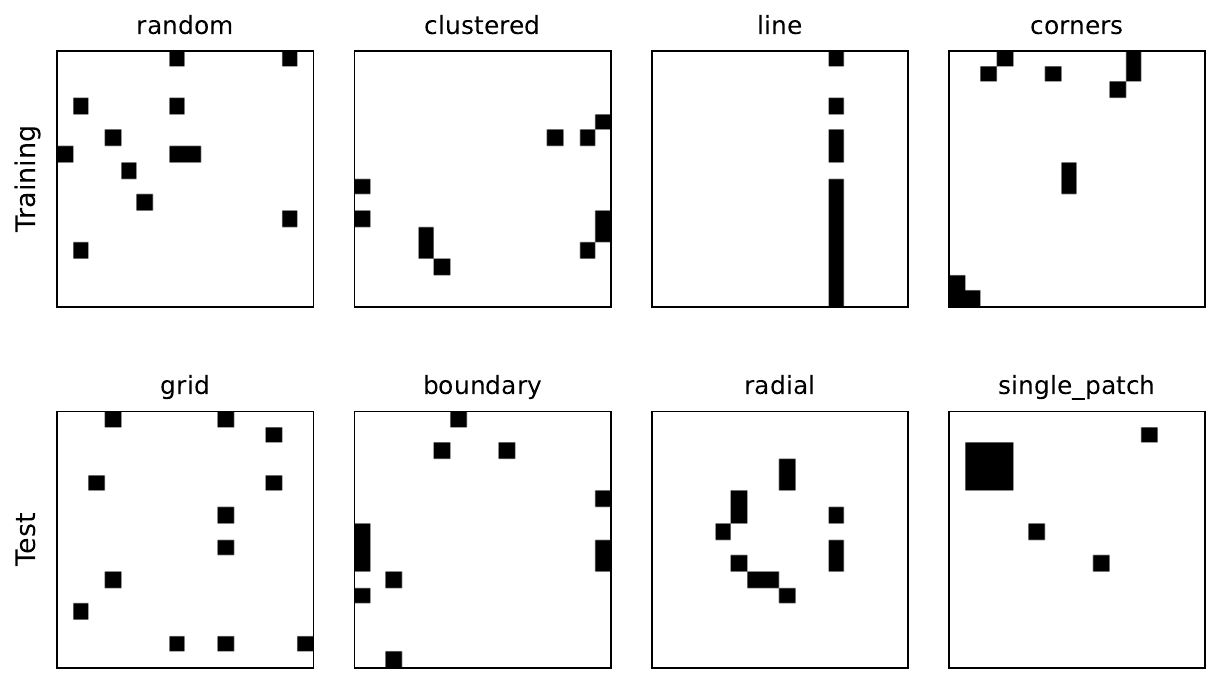}
    \caption{Sample masks of each type at a fixed 5.0\% sparsity. The top row denotes the masks used for training and the bottom row denotes the masks used during inference.}
    \label{fig:mask_gallery}
\end{figure*}

For each sample, a binary mask is generated on the low-resolution grid. Let $H_{\mathrm{lr}} \times W_{\mathrm{lr}}$ denote the LR resolution and let
\begin{equation}
n_{\mathrm{obs}} = \max\!\left(1, \left\lfloor s H_{\mathrm{lr}} W_{\mathrm{lr}} \right\rfloor\right),
\end{equation}
where $s$ is the observation fraction. The implemented mask generators are: random, clustered, line, corners, grid, boundary, radial, and single patch. In the training pipeline, $s \sim U[s_{\min}, s_{\max}]$ with $[s_{\min}, s_{\max}] = [0.01, 0.15]$ and the mask type is drawn from the first four masks listed. At evaluation time, $s$ is fixed and the mask type is drawn from the last four masks listed. The masks are visualized in Figure \ref{fig:mask_gallery}.

\subsection{Regime Ranges}
In this section we provide the parameter ranges for each of our study regimes in Table~\ref{tab:regime_ranges}. Note that for the advection--diffusion subregimes, the domains are determined by the P\'eclet number, defined as
\begin{equation}\label{eq:peclet}
    \mathrm{Pe} = \frac{UL}{\kappa} = \frac{L\sqrt{v_x^2+v_y^2}}{\kappa},
\end{equation}
where $U=\|\mathbf{v}\|$ is the characteristic speed, and $L$ is the characteristic length scale (which for $\Omega=[0,1)^2$, is just $L=1$). The reason for this distinction is that each subregime has drastically different evolution patterns.
\begin{table}[H]
\centering
\small
\caption{Regime definitions used in our experiments. The first four regimes are subregimes of advection--diffusion; the final two correspond to Klein--Gordon and Helmholtz.}
\label{tab:regime_ranges}
\begin{tabular}{l l c c c c}
\hline
Regime & PDE family & Range 1 & Range 2 & Range 3 & Forcing std. \\
\hline
\texttt{diffusion} &
Advection--diffusion &
$v_x \in [-1.0,\,1.0]$ &
$v_y \in [-1.0,\,1.0]$ &
$\kappa \in [0.02,\,0.35]$ &
$0.30$ \\

\texttt{advection} &
Advection--diffusion &
$v_x \in [-3.0,\,3.0]$ &
$v_y \in [-3.0,\,3.0]$ &
$\kappa \in [0.001,\,0.08]$ &
$0.30$ \\

\texttt{balanced} &
Advection--diffusion &
$v_x \in [-2.0,\,2.0]$ &
$v_y \in [-2.0,\,2.0]$ &
$\kappa \in [0.01,\,0.20]$ &
$0.40$ \\

\texttt{forcing} &
Advection--diffusion &
$v_x \in [-1.5,\,1.5]$ &
$v_y \in [-1.5,\,1.5]$ &
$\kappa \in [0.01,\,0.20]$ &
$1.00$ \\

\texttt{klein\_gordon} &
Klein--Gordon &
$c_x \in [0.4,\,2.8]$ &
$c_y \in [0.4,\,2.8]$ &
$m \in [0.3,\,3.5]$ &
$0.45$ \\

\texttt{helmholtz} &
Helmholtz &
$\kappa_x \in [0.03,\,0.55]$ &
$\kappa_y \in [0.03,\,0.55]$ &
$k \in [0.4,\,3.5]$ &
$0.55$ \\
\hline
\end{tabular}
\end{table}

\subsection{Forcing Term}
\label{sec:appendix_forcing}
The steady source term is represented spectrally on the periodic domain \(\Omega=[0,1)^2\). Rather than learning or optimizing a forcing value at every grid point, we parameterize \(f\) by a truncated set of low-frequency Fourier coefficients. Letting \(M_f\) denote the number of retained forcing modes, the forcing field is written as
\[
    f_{\mathbf q}(x,y) = \sum_{|k_x|<M_f} \sum_{|k_y|<M_f} \widehat f_{\mathbf q}(k_x,k_y) e^{2\pi i(k_x x+k_y y)}, \qquad \widehat f_{\mathbf q}(-k_x,-k_y) = \overline{\widehat f_{\mathbf q}(k_x,k_y)},
\]
so that \(f_{\mathbf q}\) is real-valued. Equivalently, this is a finite
two-dimensional sine-cosine expansion,
\[
    f_{\mathbf q}(x,y) = \sum_{|k_x|<M_f} \sum_{|k_y|<M_f} a_{k_x,k_y} \cos\!\left(2\pi(k_xx+k_yy)\right) + b_{k_x,k_y} \sin\!\left(2\pi(k_xx+k_yy)\right),
\]
with coefficients collected in the forcing latent vector \(\mathbf q\). In the implementation we store the coefficients in the real-FFT half-plane and recover the physical forcing field by an inverse FFT:
\[
    f_{\mathbf q} = \mathcal F^{-1}\!\left(\widehat f_{\mathbf q}\right).
\]
All Fourier coefficients outside the retained low-frequency block are set to zero, so the source is smooth and resolution-transferable. We choose \(M_f=12\), giving \(4M_f^2=576\) real latent degrees of freedom for the forcing field. This source is time-independent and enters additively in each PDE family. For example, in the evolutionary equations it appears as a steady forcing term, while in Helmholtz it contributes to the static right-hand side.

\section{Experimental Setup}

\subsection{Estimating Ensemble Members for Diffusion-Based Methods}
\label{sec:appendix_ensemble_size}

Let $\mathbf U \in \mathbb R^d$ denote the decoded random output under the conditional posterior for a fixed test instance, with posterior mean
\[
\boldsymbol{\mu} \coloneqq \mathbb E[\mathbf U \mid y,M,u_0^{\mathrm{lr}}],
\]
and covariance matrix $\Sigma$. Given $K$ independent posterior samples
$\mathbf U^{(1)},\dots,\mathbf U^{(K)} \sim p(\mathbf U \mid y,M,u_0^{\mathrm{lr}}),$ the Monte Carlo posterior mean is
\[
\widehat{\boldsymbol{\mu}}_K \coloneqq \frac{1}{K}\sum_{k=1}^K \mathbf U^{(k)}.
\]
Under the discrete $L_h^2$ norm
\[
\|\mathbf v\|_{L_h^2}^2 \coloneqq \frac{1}{d}\sum_{j=1}^d v_j^2,
\]
we have the exact identity
\begin{equation}
\mathbb E\!\left[\|\widehat{\boldsymbol{\mu}}_K-\boldsymbol{\mu}\|_{L_h^2}^2 \;\middle|\; y,M,u_0^{\mathrm{lr}}\right] = \frac{1}{K d}\operatorname{tr}(\Sigma).
\label{eq:ensemble_mc_error}
\end{equation}
Thus the Monte Carlo error of the posterior mean decays as $K^{-1/2}$. A good choice of $K$ is therefore the smallest value such that
\begin{equation}
\frac{1}{K d}\operatorname{tr}(\Sigma) \le \varepsilon_{\mathrm{ens}}^2,
\qquad\text{i.e.}\qquad
K \ge \frac{\operatorname{tr}(\Sigma)}{d\,\varepsilon_{\mathrm{ens}}^2},
\label{eq:ensemble_size_absolute}
\end{equation}
for a prescribed posterior-mean tolerance $\varepsilon_{\mathrm{ens}}$. Equivalently, if one requires the Monte Carlo standard error to be at most an $\eta$-fraction of the posterior standard deviation, then $K \ge \eta^{-2}$. In practice, $\operatorname{tr}(\Sigma)$ is unknown and is estimated from a pilot ensemble, after which \eqref{eq:ensemble_size_absolute} is enforced conservatively over evaluation strata.

\begin{figure}[h!]
    \centering
    \includegraphics[width=0.5\linewidth]{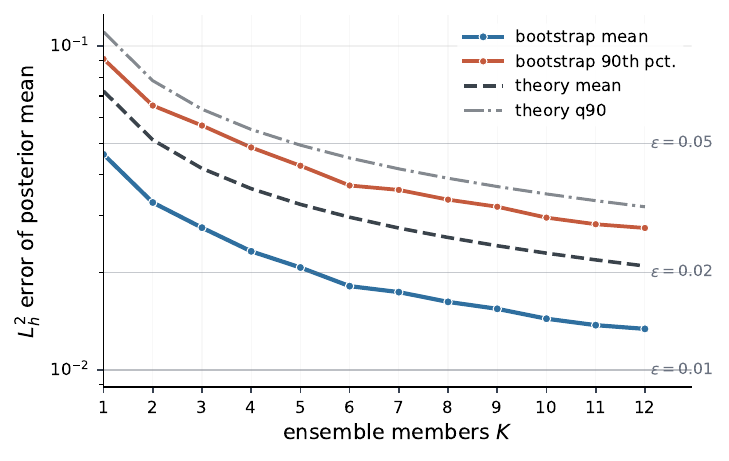}
    \caption{Monte Carlo convergence of the LatentPDE posterior mean as a function of the number of diffusion ensemble members \(K\). Empirical errors are computed by subsampling posterior samples and comparing the resulting ensemble mean to the full pilot-ensemble mean, while theoretical curves use the estimate \(\operatorname{tr}(\widehat{\Sigma})/(dK)\) from the posterior covariance, as described by Equation~\ref{eq:ensemble_mc_error}.}
    \label{fig:ensemble_mean}
\end{figure}

An empirical study of this can be found in Figure~\ref{fig:ensemble_mean}. We have chosen $K=12$ for as our number of ensemble members as that is where the theoretical posterior mean reaches an error of $\epsilon=0.02$.

\subsection{Partitioning the Dataset}
\label{sec:appendix_train_test}

For each regime in Table~\ref{tab:regime_ranges}, we generate independent train, validation, and test sets using the same parameter ranges. Each regime provides 5000 training examples, alongside 500 validation and 500 test example. To prevent data leakage, we ensure that no samples across the different splits share the same exact initial conditions, coefficient draws, forcing functions, or masks. As such, all samples in the train, validation, and test sets are completely independent from each regime-specific distribution, except for the intentional shift in mask distribution between training and evaluation.

For each regime, we evaluate both \textsc{LatentPDE-MAP} and \textsc{LatentPDE-Enc} on the regime's validation split, averaging RMSE over all four evaluation masks. We then select the variant with lower average validation error and freeze that choice before touching the held-out test set. The selected variant is used for the entirety of the corresponding test split for that regime. 

\subsection{Evaluation Metrics}
\label{sec:appendix_eval_metrics}
Let $u \in \mathbb R^{H\times W}$ denote the ground-truth field on a discrete grid, let $\hat u \in \mathbb R^{H\times W}$ denote a point prediction, and let
\[
\{u^{(s)}\}_{s=1}^S, \qquad u^{(s)} \in \mathbb R^{H\times W},
\]
denote an ensemble of posterior samples when available. We write $d \coloneqq HW$ for the number of spatial degrees of freedom and identify each field with its vectorization in $\mathbb R^d$ when convenient.

\paragraph{Root Mean Square Error (RMSE).}
For a point prediction $\hat u$ and truth $u$, the per-instance RMSE is
\begin{equation}
\mathrm{RMSE}(\hat u,u)
\coloneqq
\left(
\frac{1}{d}\sum_{j=1}^d (\hat u_j-u_j)^2
\right)^{1/2}.
\end{equation}
Equivalently, RMSE is the discrete $L^2$ error normalized by the number of grid points. Over a test set $\{(\hat u^{(i)},u^{(i)})\}_{i=1}^N$, we report the empirical mean
\begin{equation}
\overline{\mathrm{RMSE}}
=
\frac{1}{N}\sum_{i=1}^N \mathrm{RMSE}(\hat u^{(i)},u^{(i)}).
\end{equation}

\paragraph{Radially Averaged PSD Log Error.}
To assess scale-by-scale spectral fidelity, we compare radially averaged power spectral densities (PSDs). For a field $u$, let
\[
u^\circ \coloneqq u - \bar u,
\qquad
\bar u \coloneqq \frac{1}{d}\sum_{j=1}^d u_j,
\]
and let $\widehat{u^\circ}(\boldsymbol{\xi})$ denote the discrete Fourier transform at frequency $\boldsymbol{\xi}$. The two-dimensional periodogram is
\begin{equation}
P_u^{2\mathrm{D}}(\boldsymbol{\xi}) \coloneqq
\frac{|\widehat{u^\circ}(\boldsymbol{\xi})|^2}{HW}.
\end{equation}
For integer radii $k=1,\dots,K_{\max}$ with $K_{\max}=\lfloor \min(H,W)/2 \rfloor$, define the annulus
\begin{equation}
A_k\coloneqq \left\{ \boldsymbol{\xi} :
k-\tfrac12 \le \|\boldsymbol{\xi}\|_2 < k+\tfrac12
\right\},
\end{equation}
and the radially averaged PSD
\begin{equation}
P_u(k)\coloneqq
\frac{1}{|A_k|}
\sum_{\boldsymbol{\xi}\in A_k} P_u^{2\mathrm{D}}(\boldsymbol{\xi}).
\end{equation}
The PSD log error between a prediction $\hat u$ and truth $u$ is then
\begin{equation}
\mathrm{PSDLogErr}(\hat u,u) \coloneqq
\left(
\frac{1}{K_{\max}}
\sum_{k=1}^{K_{\max}}
\left[
\log_{10}\!\bigl(\max(P_{\hat u}(k),\varepsilon)\bigr)
-
\log_{10}\!\bigl(\max(P_u(k),\varepsilon)\bigr)
\right]^2
\right)^{1/2},
\end{equation}
where $\varepsilon>0$ is a small floor used for numerical stability. This metric is insensitive to phase alignment and instead measures whether the prediction places the correct amount of energy at each spatial scale. We use it to complement RMSE by distinguishing physical-space accuracy from spectral consistency.

\paragraph{Continuous Ranked Probability Score (CRPS).}
For probabilistic predictions, we evaluate the ensemble forecast using the empirical CRPS applied pointwise and then averaged over the grid. For each spatial coordinate $j \in \{1,\dots,d\}$, let $\{u^{(1)}_j,\dots,u^{(S)}_j\}$ be the ensemble of scalar predictions for the true value $u_j$. The empirical CRPS at coordinate $j$ is
\begin{equation}
\mathrm{CRPS}_j
=
\frac{1}{S}\sum_{s=1}^S |u_j^{(s)}-u_j|
-
\frac{1}{2S(S-1)}
\sum_{\substack{s,t=1\\ s\neq t}}^S
|u_j^{(s)}-u_j^{(t)}|.
\end{equation}
The field-level CRPS is the spatial average
\begin{equation}
\mathrm{CRPS}\bigl(\{u^{(s)}\}_{s=1}^S,u\bigr)
=
\frac{1}{d}\sum_{j=1}^d \mathrm{CRPS}_j.
\end{equation}

\subsection{Comparison Methods}
\label{sec:appendix_comparison_methods}
\paragraph{3D-Var.}
In this section we briefly describe 3D-Var for those unfamiliar with data assimilation techniques. We note first that we operate directly on the high-resolution output field and require a high-resolution initial condition $u_0^{\mathrm{hr}}$ (approximated by bilinear upsampling of $u_0^{\mathrm{lr}}$ at test time), unlike LatentPDE which conditions only on $u_0^{\mathrm{lr}}$.
\begin{equation}
    J(\mathbf{x}) = \tfrac{1}{2}(\mathbf{x} - \mathbf{x}_b)^\top B^{-1}(\mathbf{x} - \mathbf{x}_b) + \tfrac{1}{2}(y - \mathcal{H}(\mathbf{x}))^\top R^{-1}(y - \mathcal{H}(\mathbf{x})),
\end{equation}
where $\mathbf{x}_b$ is a background state, $B$ is the background error covariance, $R = \sigma_{\mathrm{obs}}^2 I$ is the observation error covariance, and $\mathcal{H}(\mathbf{x}) = M \odot \mathcal{D}_s(\mathbf{x})$ is the linear observation operator composed of spatial downsampling $\mathcal{D}_s$ and masking by $M$. We initialize the background $\mathbf{x}_b$ using either MAP or encoder prediction, consistent with LatentPDE.

Since $B$ is $n \times n$ and infeasible to form explicitly on the HR grid, we parameterize $B^{-1}$ implicitly through the diffusion-operator construction of~\citet{weaver2001correlation}:
\begin{equation}
    B^{-1} \approx \sigma_b^{-2}\bigl(I - L^2 \nabla^2\bigr)^p,
\end{equation}
where $\sigma_b$ is the background standard deviation, $L$ is a spatial correlation length, and $\nabla^2$ is the discrete five-point Laplacian with Neumann boundary conditions. The power $p$ controls the smoothness of the implied correlation function. We use $p = 2$, which approximates a Mat\'ern-3/2 kernel. 

\begin{figure}[h!]
    \centering
    \includegraphics[width=0.5\linewidth]{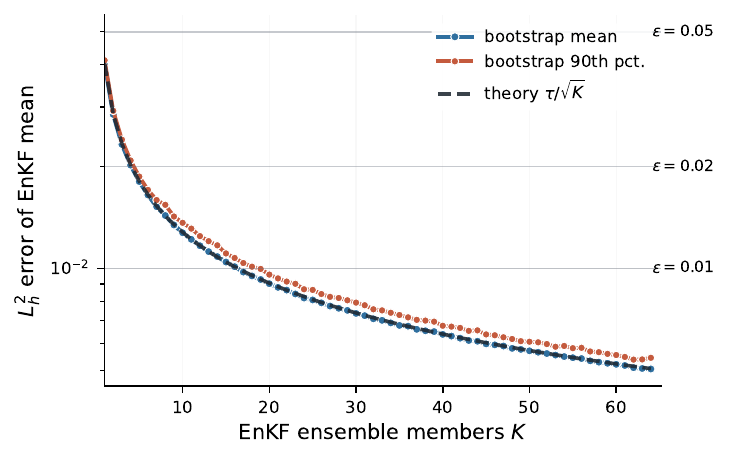}
    \caption{Monte Carlo convergence of the EnKF analysis mean as a function of the number of ensemble members \(K\). We choose $K=16$ for where the error $\epsilon$ drops below 0.01. For a more detailed explanation of how this number was derived, refer to Section~\ref{sec:appendix_ensemble_size}.}
    \label{fig:enkf_ensemble}
\end{figure}

\paragraph{Ensemble Kalman Filtering (EnKF).}
In contrast to 3D-Var's fixed parametric $B$, the Ensemble Kalman Filter~\citep{evensen1994sequential} estimates a flow-dependent background covariance from an ensemble of perturbed states. We initialize the background $\mathbf{x}_b \in \mathbb{R}^{H \times W}$ using MAP or encoder prediction, matching the 3D-Var setup. An ensemble of $N_e = 16$ (refer to Figure~\ref{fig:enkf_ensemble} for a plot of how this was derived) members is constructed by adding spatially-smoothed, observation-masked Gaussian noise:
\begin{equation}
    \mathbf{x}_b^{(i)} = \mathbf{x}_b + \sigma_e\, S\!\left[(1 - \tfrac{1}{2}M_{\mathrm{hr}}) \odot \boldsymbol{\xi}^{(i)}\right], \quad \boldsymbol{\xi}^{(i)} \sim \mathcal{N}(0,I),
\end{equation}
where $S[\cdot]$ denotes repeated average pooling for spatial smoothing and $M_\mathrm{hr}$ is the HR-upsampled mask. The analysis update is then applied for $C = 2$ cycles. In each cycle, the ensemble members are passed through the observation operator $\mathcal{H}(\mathbf{x}) = M \odot \mathcal{D}_s(\mathbf{x})$ to form the predicted observation anomalies $A_y \in \mathbb{R}^{N_e \times p}$ and state anomalies $A_x \in \mathbb{R}^{N_e \times n}$.  The ensemble Kalman gain is applied with perturbed observations~\cite{burgers1998}:
\begin{equation}
    \mathbf{x}^{(i)} \leftarrow \mathbf{x}^{(i)}
    + \frac{1}{N_e - 1} A_x^\top A_y
    \left(\frac{A_y^\top A_y}{N_e - 1} + \sigma_{\mathrm{obs}}^2 I\right)^{-1}
    \!\!\bigl(y^{(i)} - \mathcal{H}(\mathbf{x}^{(i)})\bigr),
\end{equation}
where $y^{(i)} = y + \sigma_{\mathrm{obs}}\,\boldsymbol{\eta}^{(i)}$,   $\boldsymbol{\eta}^{(i)} \sim \mathcal{N}(0,I)$, are perturbed observations. The analysis field is the ensemble mean, post-smoothed and corrected at observed locations by the direct residual $M \odot (y - \mathcal{D}_s(\bar{\mathbf{x}}))$.

\paragraph{Family-specific residuals for PINNs.}
For completeness, the PDE residuals appearing in $\mathcal{L}_{\mathrm{pde}}$ are given as follows:
\begin{align*}
    r_{\eta,\mathsf{AD}} &=
    \partial_{t}u_{\eta}
    + v_{x}\partial_{x}u_{\eta}
    + v_{y}\partial_{y}u_{\eta}
    - \kappa(\partial_{xx}u_{\eta} + \partial_{yy}u_{\eta})
    - f_{\zeta}, \\
    r_{\eta,\mathsf{KG}} &=
    \partial_{tt}u_{\eta}
    - c_{x}^{2}\partial_{xx}u_{\eta}
    - c_{y}^{2}\partial_{yy}u_{\eta}
    + m^{2}u_{\eta}
    - f_{\zeta}, \\
    r_{\eta,\mathsf{H}} &=
    -\kappa_{x}\partial_{xx}u_{\eta}
    -\kappa_{y}\partial_{yy}u_{\eta}
    + k^{2} u_{\eta}
    - (u_{0}^{\mathrm{hr}} + f_{\zeta}).
\end{align*}

\section{Impact of Sampling Perturbations on Reconstruction Accuracy}
\begin{figure*}[h!]
    \centering
    \includegraphics[width=.8\textwidth]{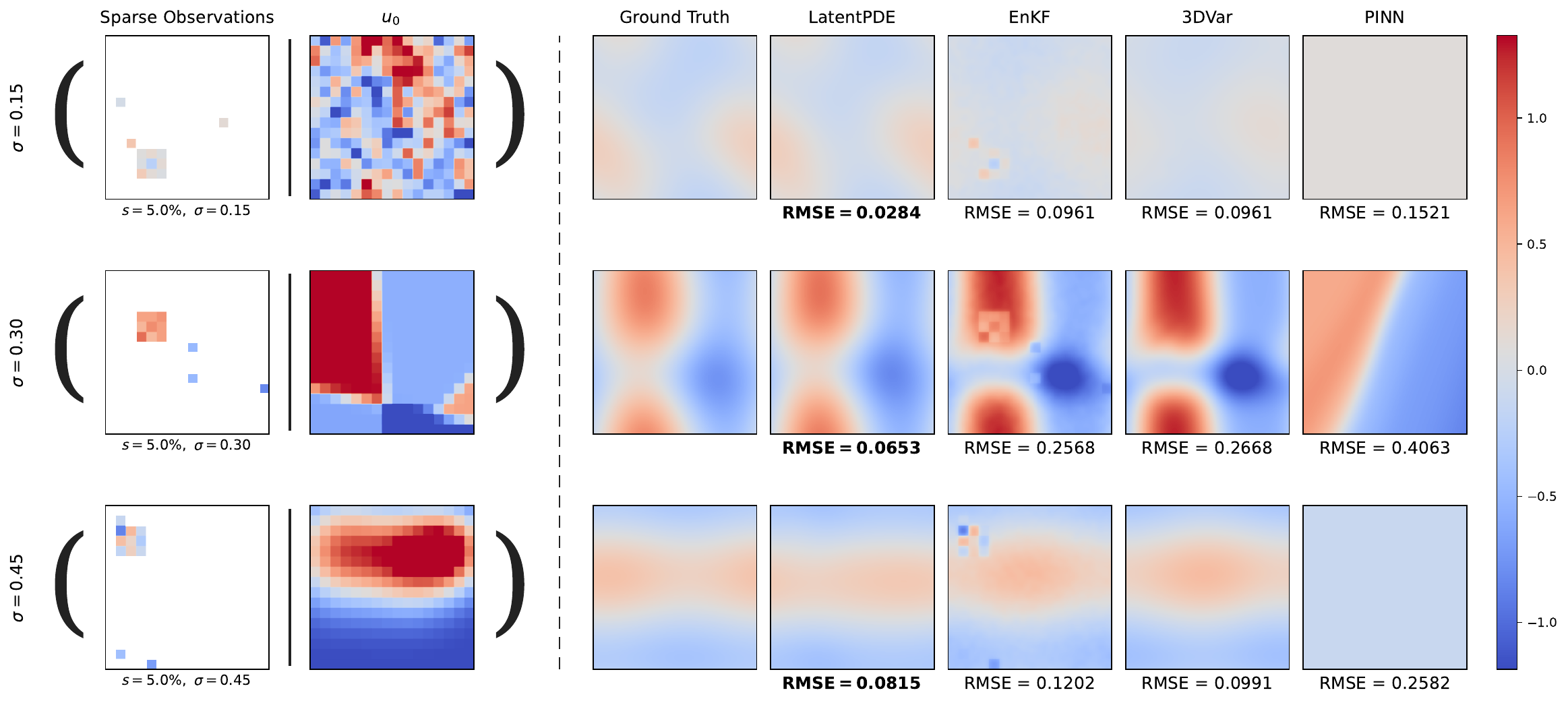}
    \caption{Reconstructions of the advection--diffusion equation under increasing observation noise ($\sigma$) at a fixed 5.0\% sparsity.}
    \label{fig:noise_advdiff}
\end{figure*}

\begin{figure*}[h!]
    \centering
    \includegraphics[width=.8\textwidth]{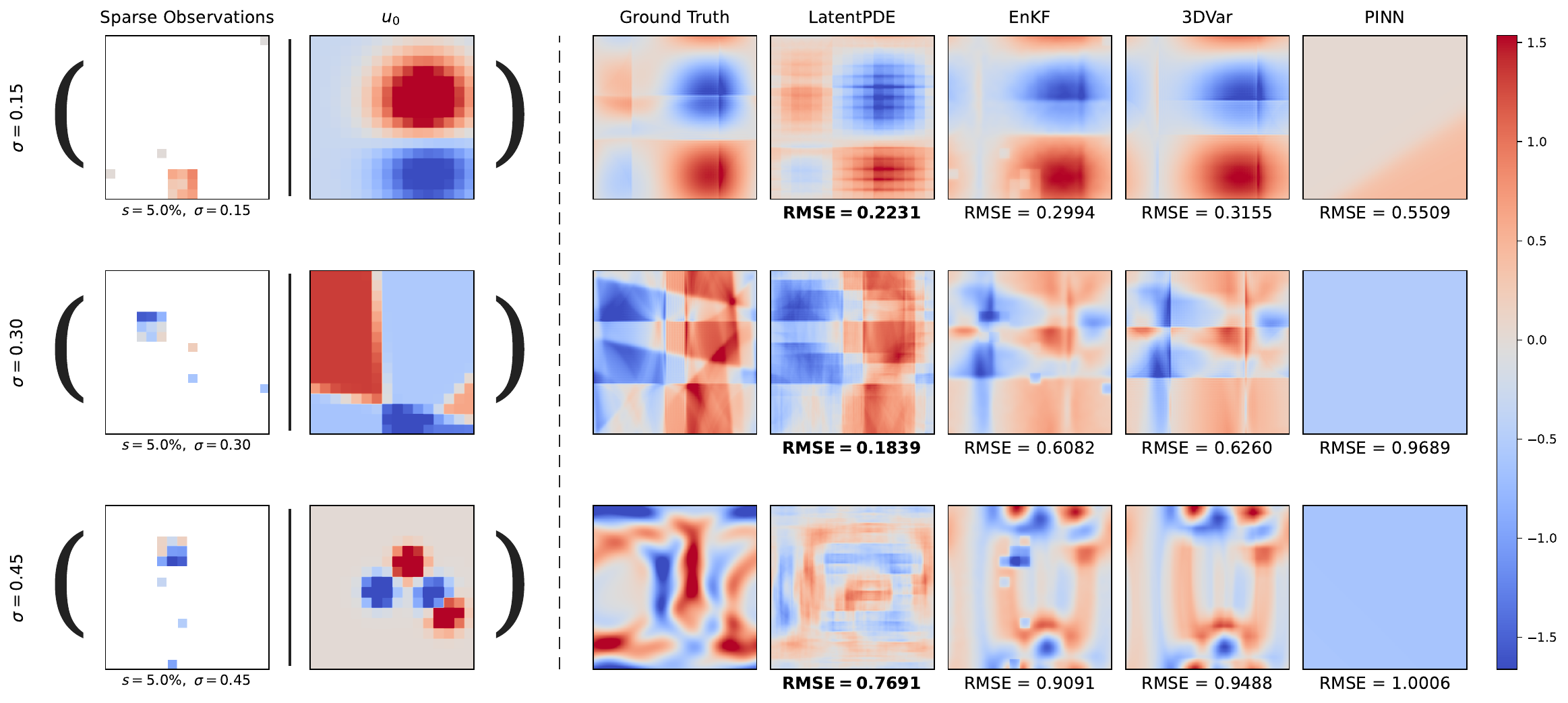}
    \caption{Reconstructions of the Klein--Gordon equation under increasing observation noise ($\sigma$) at a fixed 5.0\% sparsity}
    \label{fig:noise_kg}
\end{figure*}

\begin{figure*}[h!]
    \centering
    \includegraphics[width=.8\textwidth]{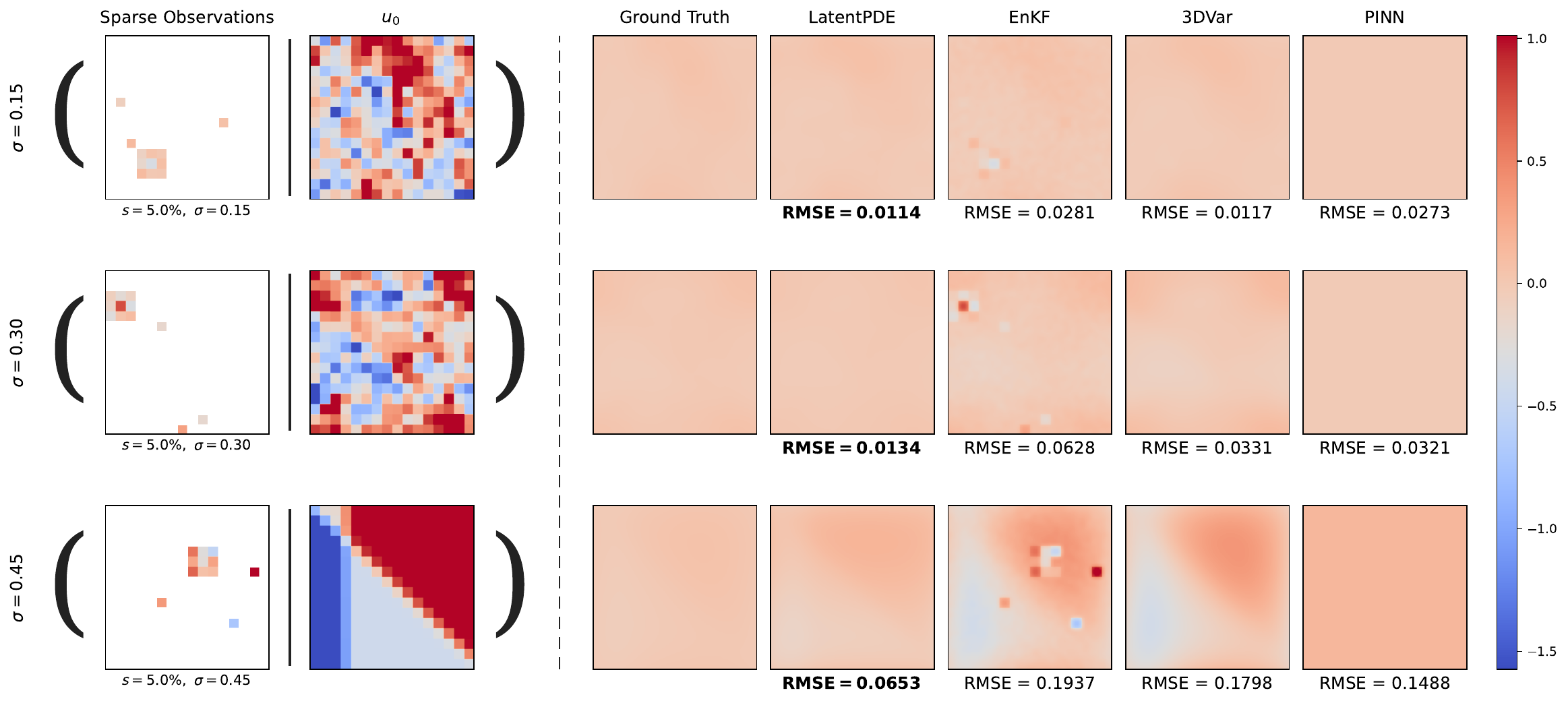}
    \caption{Reconstructions of the Helmholtz equation under increasing observation noise ($\sigma$) at a fixed 5.0\% sparsity}
    \label{fig:noise_hemlholtz}
\end{figure*}

In order to investigate the reconstruction quality of LatentPDE under varying levels of injected noise, Figures \ref{fig:noise_advdiff} through \ref{fig:noise_hemlholtz} below provide a qualitative comparison at observational noise levels of $0.15$, $0.30$, and $0.45$. While EnKF suffers from localized artifacts around known observation points, 3D-Var and PINNs exhibit severe over-smoothing at higher noise levels ($\sigma = 0.45$). LatentPDE however maintains accurate global field reconstructions. These trends in noise robustness are consistently observed across our Klein--Gordon and Helmholtz evaluations as well.

\clearpage
\section{Robustness to Mask Variations}
\begin{figure*}[h!]
    \centering
    \includegraphics[width=.8\textwidth]{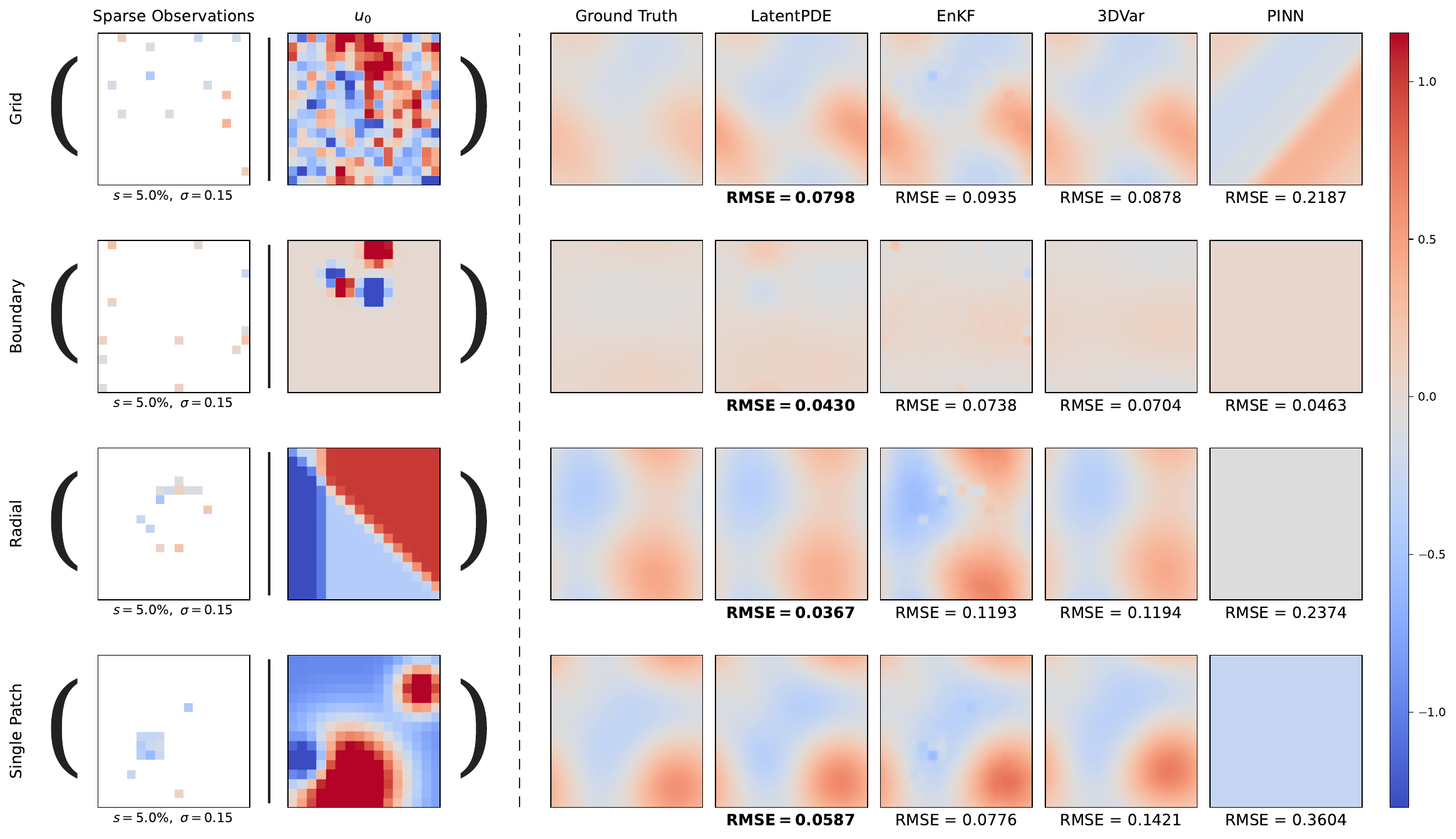}
    \caption{Reconstructions of the advection--diffusion equation under diverse mask configurations (grid, boundary, radial, and single patch). For all realizations here, $\sigma$ is fixed at 0.15 and sparsity at 5.0\%.}
    \label{fig:masks_advdiff}
\end{figure*}

\begin{figure*}[h!]
    \centering
    \includegraphics[width=.8\textwidth]{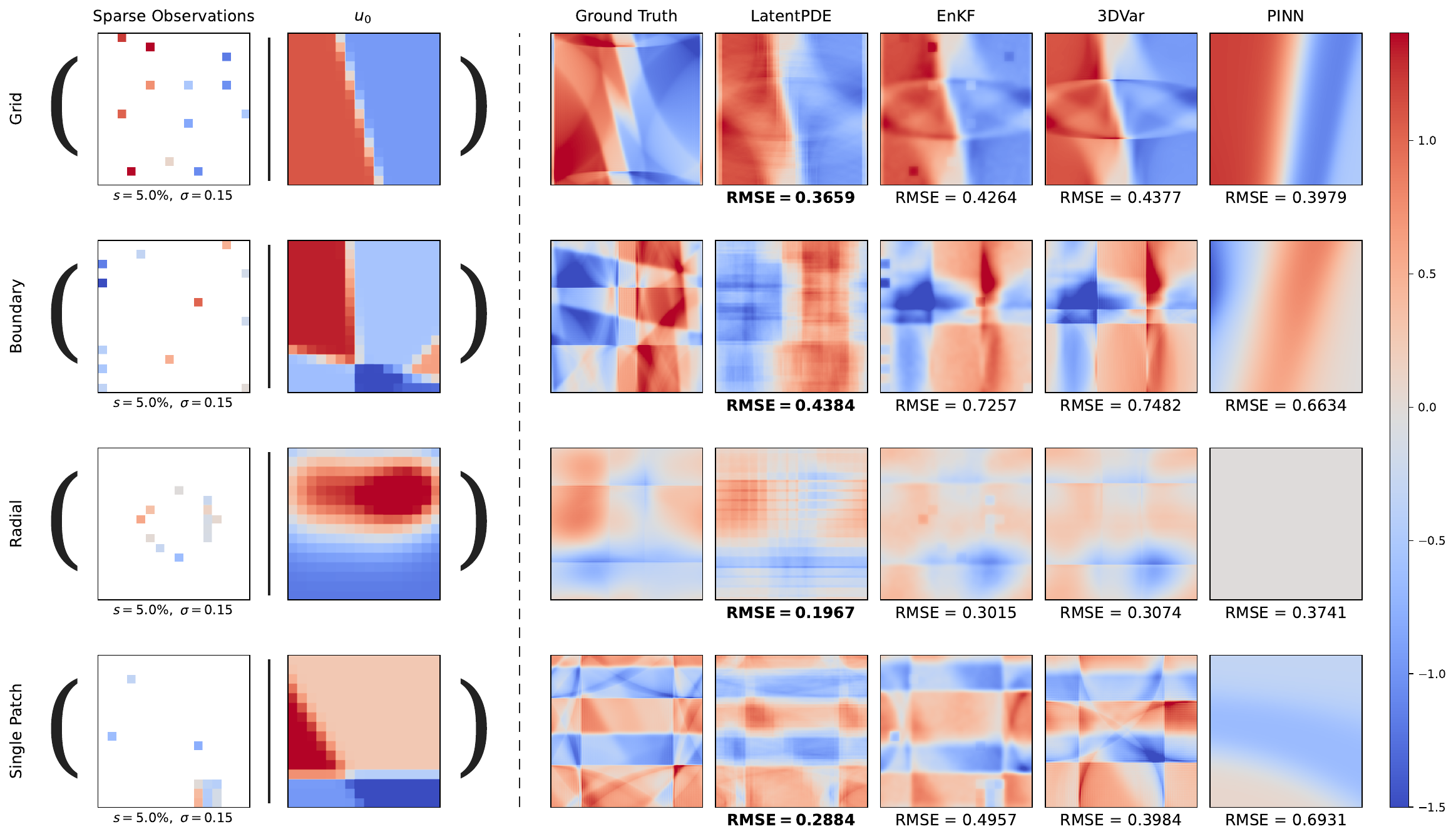}
    \caption{Reconstructions of the Klein--Gordon equation under diverse mask configurations (grid, boundary, radial, and single patch). For all realizations here, $\sigma$ is fixed at 0.15 and sparsity at 5.0\%.}
    \label{fig:masks_kg}
\end{figure*}

\begin{figure*}[h!]
    \centering
    \includegraphics[width=.8\textwidth]{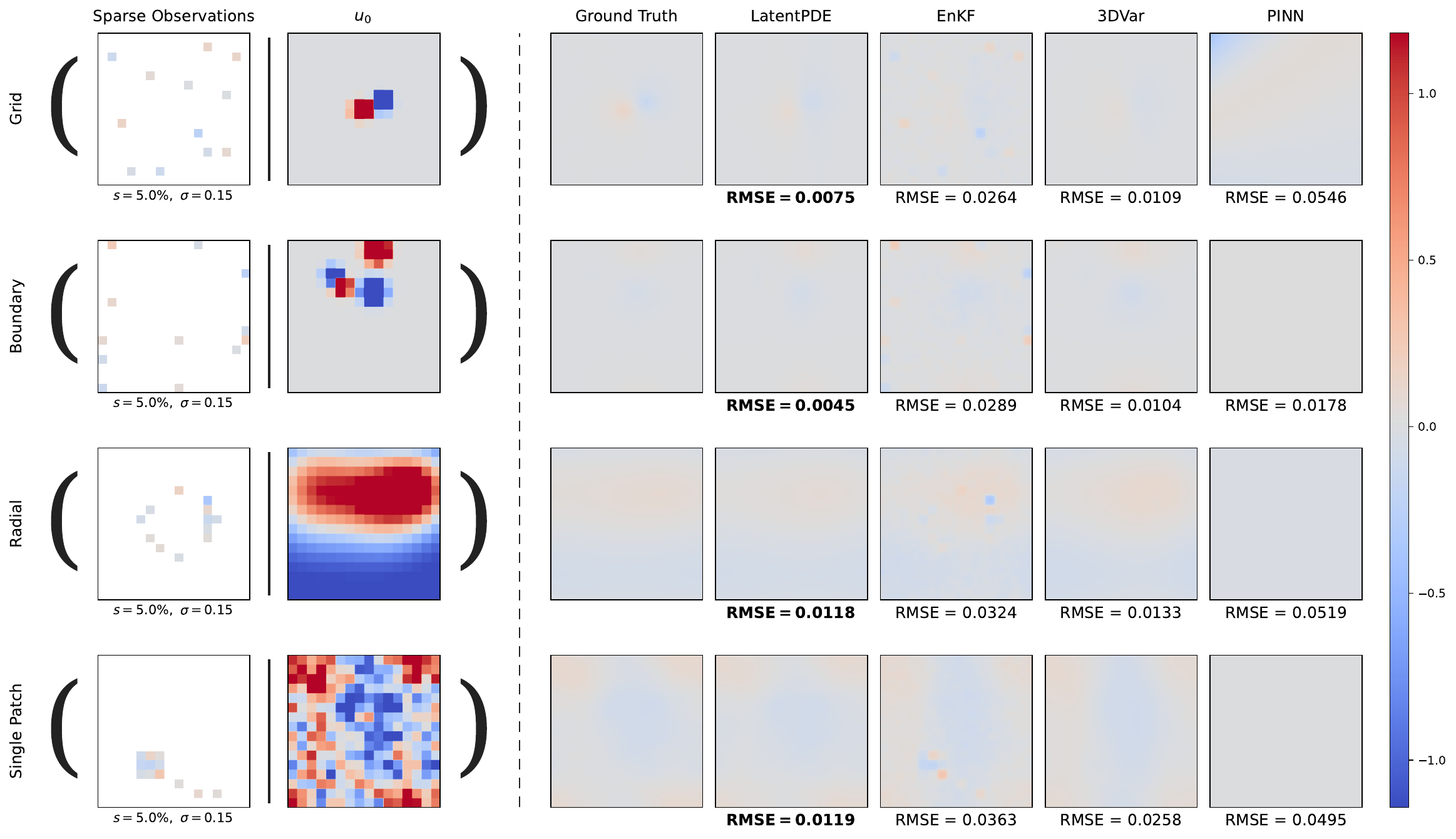}
    \caption{Reconstructions of the Helmholtz equation under diverse mask configurations (grid, boundary, radial, and single patch). For all realizations here, $\sigma$ is fixed at 0.15 and sparsity at 5.0\%.}
    \label{fig:masks_helmholtz}
\end{figure*}

Additionally, we explore the robustness of the framework under diverse spatial patterns of known data. Qualitatively, the results show that LatentPDE is learning to preserve the structural fidelity of samples during spatial reconstruction. This is perhaps most evident in Figure \ref{fig:masks_kg}, where the governing dynamics produce sharp and abrupt patterns. LatentPDE is still able to capture these global structures under severe observational sparsity and convert them to a higher resolution. In both the advection--diffusion (Figure \ref{fig:masks_advdiff}) and Helmholtz (Figure \ref{fig:masks_helmholtz}) equations, LatentPDE shows an impressive ability to extrapolate smooth gradients when given highly constrained spatial masks, whereas other methods overfit or are otherwise unable to exploit the known observations. This trend suggests that LatentPDE is effectively learning a strong global prior of the underlying physics within its latent space.

\clearpage
\section{Robustness to Varying Levels of Sparsity}
We similarly evaluate the effect of increasing sparsity levels on the reconstruction quality. Across the board in all PDEs, looking at the single patch mask (the most challenging for reconstruction), LatentPDE achieves the lowest RMSE from 1\% to 5\% sparsity.

\begin{figure*}[h!]
    \centering
    \includegraphics[width=.8\textwidth]{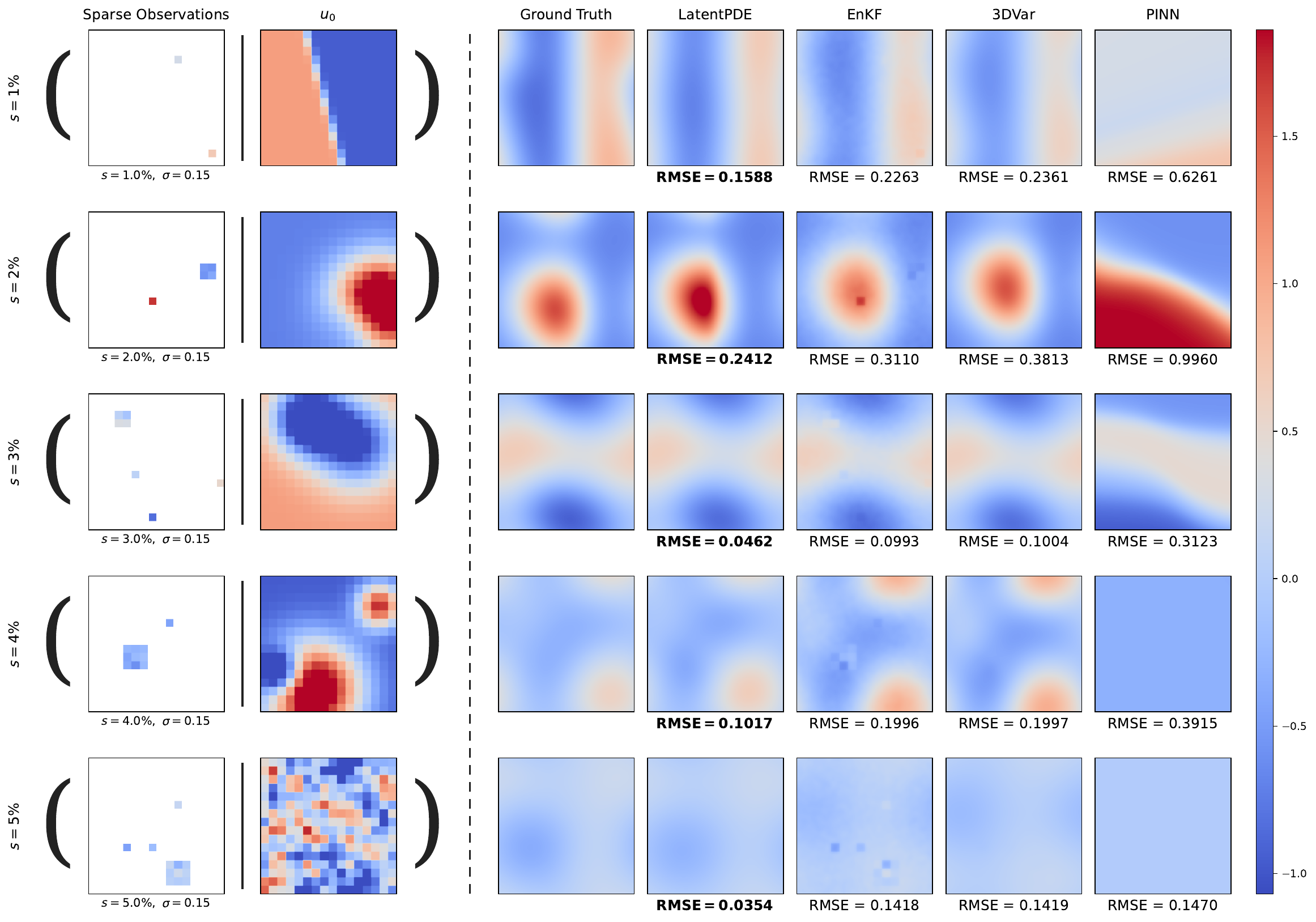}
    \caption{Qualitative comparison of reconstructions for advection--diffusion at 5 levels of sparsity (from 1\% to 5\%) for the single-patch mask. Noise ($\sigma$) is fixed at 0.15.}
    \label{fig:sparsity_advdiff}
\end{figure*}

\begin{figure*}[h!]
    \centering
    \includegraphics[width=.8\textwidth]{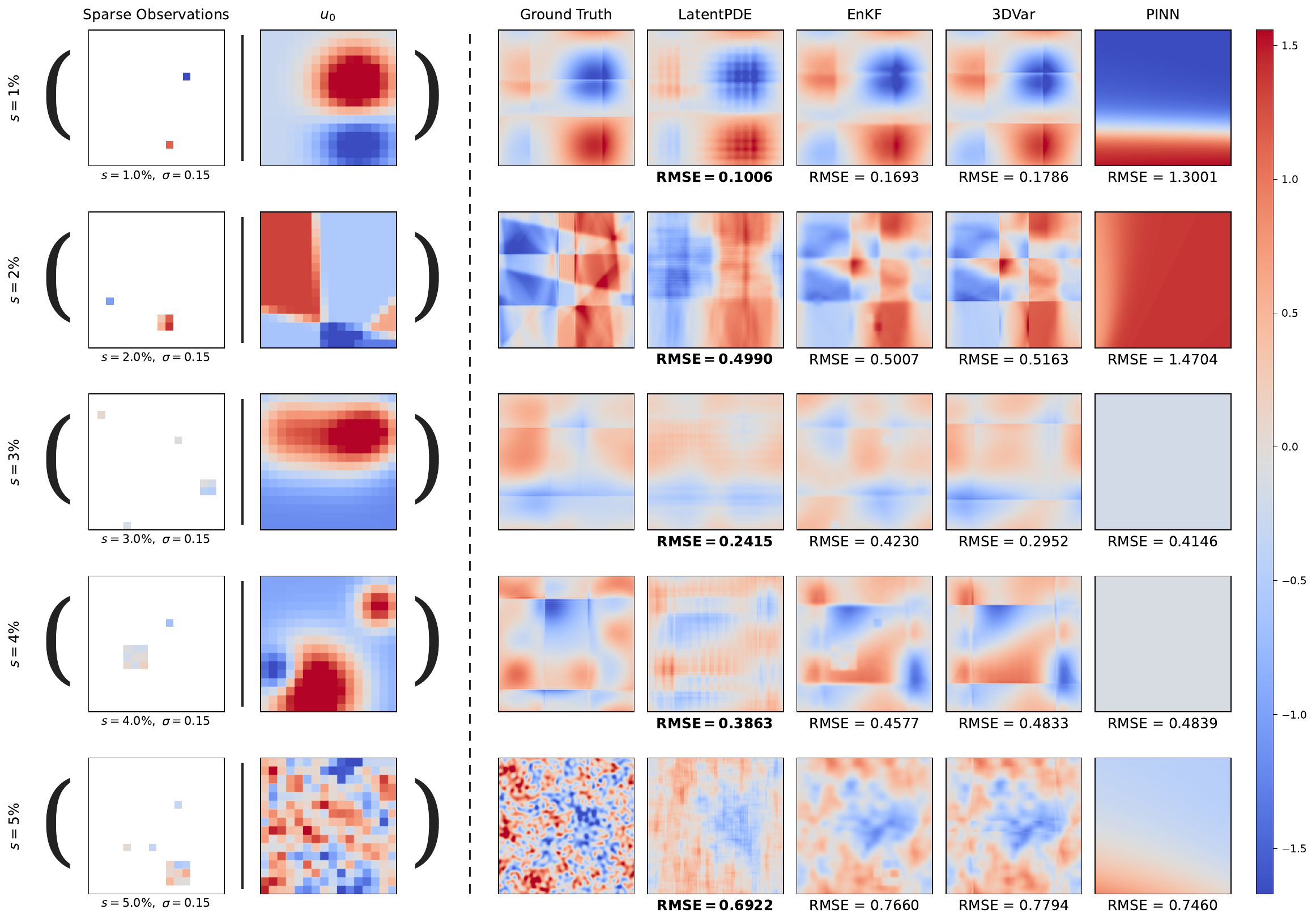}
    \caption{Qualitative comparison of reconstructions for Klein--Gordon at 5 levels of sparsity (from 1\% to 5\%) for the single-patch mask. Noise ($\sigma$) is fixed at 0.15.}
    \label{fig:sparsity_kg}
\end{figure*}

\begin{figure*}[h!]
    \centering
    \includegraphics[width=.8\textwidth]{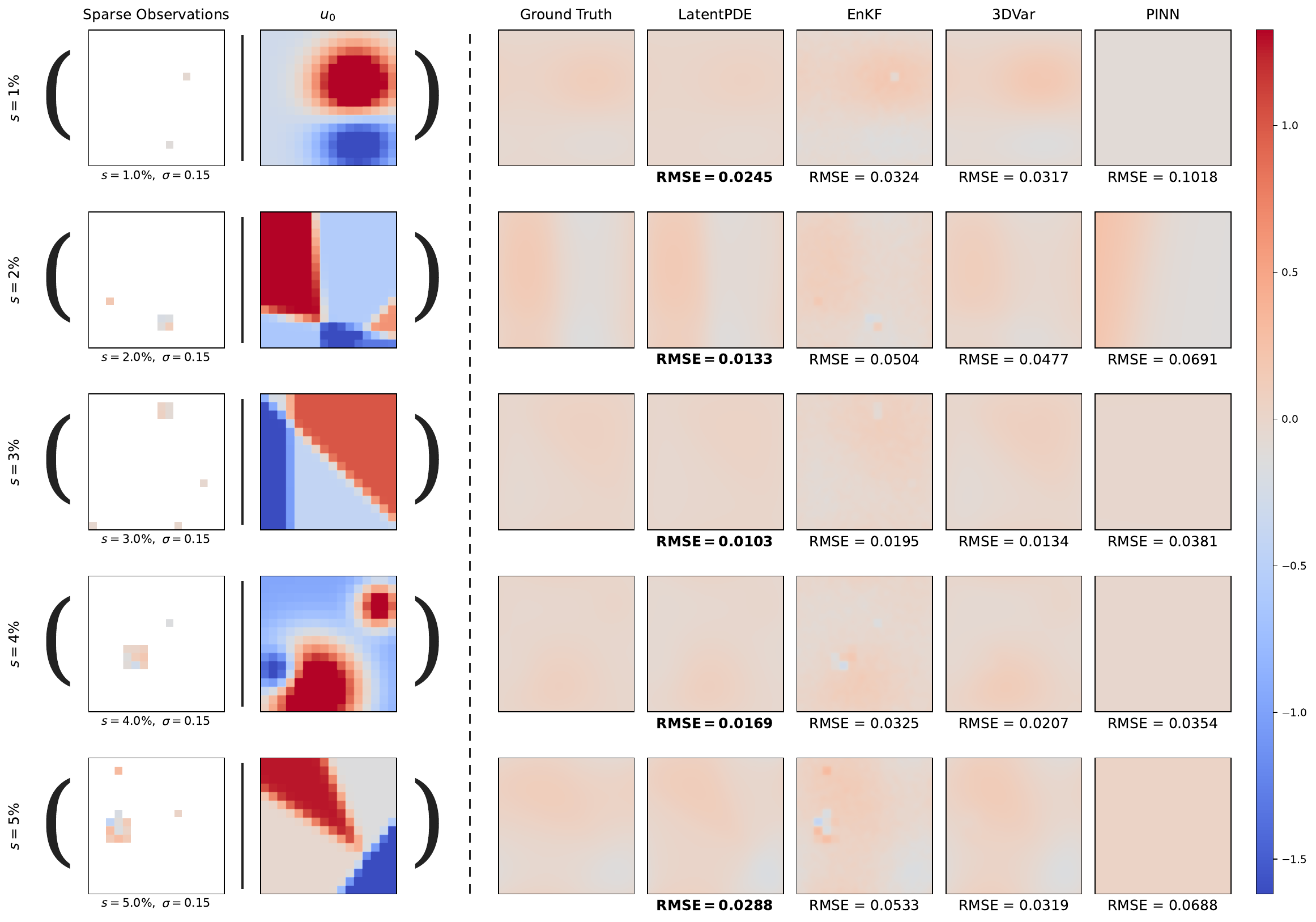}
    \caption{Qualitative comparison of reconstructions for Helmholtz at 5 levels of sparsity (from 1\% to 5\%) for the single-patch mask. Noise ($\sigma$) is fixed at 0.15.}
    \label{fig:sparsity_helmholtz}
\end{figure*}

\clearpage
\section{Random Samples from Posterior Distribution}
For thoroughness, we visualize a few reconstruction samples from each of the PDE families to show the generative nature of our model. In practicality, when we present our results in the main section, we produce a total of 12 generations for each run and average them. The justification for this is further elaborated in Section \ref{sec:baselines}.

\begin{figure*}[h!]
    \centering
    \includegraphics[width=.8\textwidth]{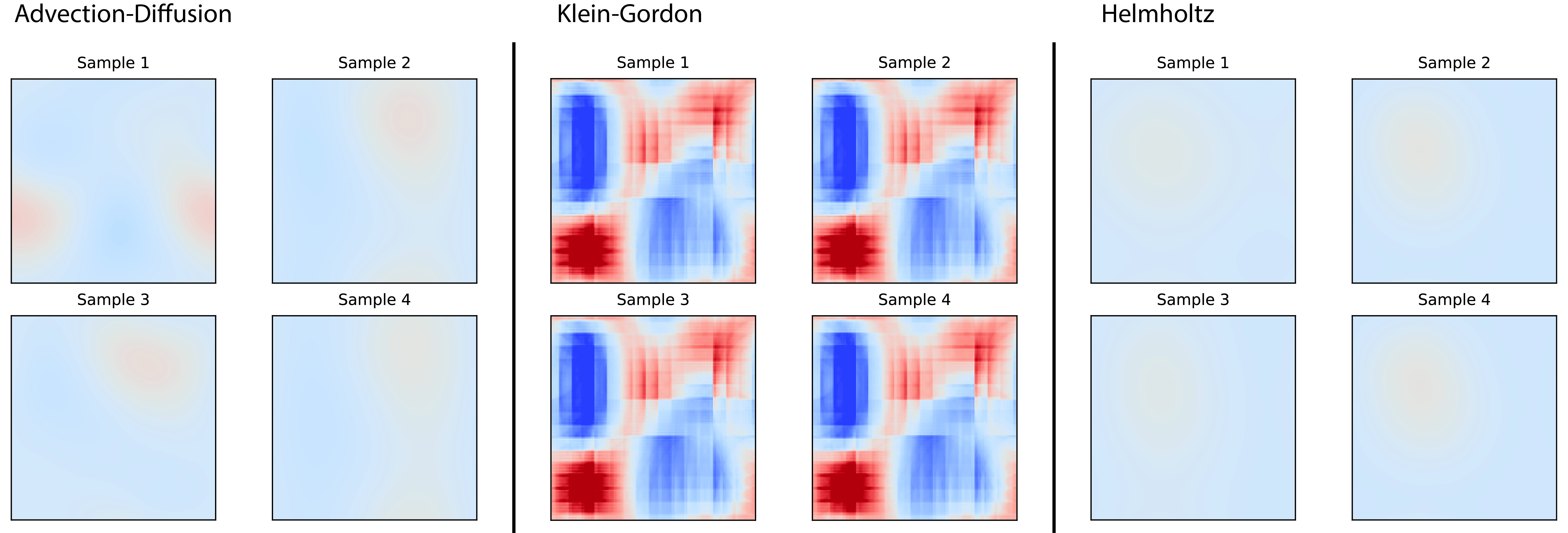}
    \caption{Random realizations generated by LatentPDE for the different equation scenarios at $\sigma=0.15$ and $s=5.0\%$. This visualization is meant to show a subset of the ensemble average that we use to report our results.}
    \label{fig:posterior_samples}
\end{figure*}

\section{Ablation Study}
\label{sec:appendix_inference_ablation}

Our ablation study evaluates three inference-time modifications to the standard diffusion model setup. We follow standard practice and compare variants using RMSE under the same downstream inference protocol.

All ablation studies are run on $5000$ training samples per subregime of advection--diffusion data with $s=5\%$ and $\sigma=0.15$. In our default configuration, we use $K=12$ posterior samples, an $80$-step MAP initialization for \textsc{LatentPDE-MAP}, reconstruction guidance enabled with scale $80$, three guidance steps per eligible reverse-diffusion step, and final latent refinement enabled for $20$ Adam steps. 

\paragraph{MAP budget.}
The MAP-budget ablation asks whether \textsc{LatentPDE-MAP} needs a well-converged test-time MAP initialization. Reducing the initialization from $80$ to $20$ Adam steps did not hurt performance in this run; it improved RMSE by $9.9\%$. The intermediate $40$-step setting worsened RMSE by $8.9\%$, suggesting that the dependence on MAP budget is not monotone in this run. We suspect that this is due to the stochasticity of the posterior sampling; having a good initialization does not have a large impact on the final reconstruction.

\begin{table}[h!]
    \centering
    \begin{tabular}{c c c}
    \hline
    MAP init. steps & RMSE & Change \\
    \hline
    80 & $0.1223$ & -- \\
    40 & $0.1332$ & $+8.9\%$ \\
    20 & $0.1103$ & $-9.9\%$ \\
    \hline
    \end{tabular}
    \caption{MAP-initialization budget ablation for \textsc{LatentPDE-MAP}.}
    \label{tab:ablation_map_budget}
\end{table}

\clearpage
\paragraph{Reconstruction guidance.}
Removing reconstruction guidance has a modest effect compared with the other inference knobs. \textsc{LatentPDE-Enc} worsens by $3.9\%$, while \textsc{LatentPDE-MAP} changes only slightly and improves by $1.1\%$ in mean RMSE. This suggests that, for these advection--diffusion runs, guidance is helpful for the encoder branch but is not the dominant source of accuracy.

\begin{table}[h!]
    \centering
    \begin{tabular}{c c c c}
    \hline
    Branch & $\gamma_g$ & RMSE & Change \\
    \hline
                & $80$ & $0.1223$ & -- \\
       \textsc{LatentPDE-MAP}  & $40$ & $0.1216$ & $-0.6\%$ \\
                             & $0$ & $0.1210$ & $-1.1\%$ \\
    \midrule \addlinespace                             
                      & $80$ & $0.1033$ & -- \\
    \textsc{LatentPDE-Enc} & $40$ & $0.1045$ & $+1.2\%$ \\
                        & $0$ & $0.1073$ & $+3.9\%$ \\
    \hline
    \end{tabular}
    \caption{Reconstruction-guidance ablation. We vary the guidance scale $\gamma_g$ while keeping the guidance step size fixed.}
    \label{tab:ablation_guidance}
\end{table}

\paragraph{Final refinement.}
The effect of final refinement is branch-dependent. Removing refinement substantially hurts \textsc{LatentPDE-Enc}, increasing RMSE by $30.4\%$, which indicates that the encoder-initialized sampler relies on the final local observation-correction step. In contrast, \textsc{LatentPDE-MAP} improves by $16.1\%$ without refinement, suggesting that the default refinement can over-correct MAP-initialized samples under sparse noisy observations in this setting.

\begin{table}[h!]
    \centering
    \begin{tabular}{c c c c}
    \hline
    Branch & $\lambda_{\mathrm{ref}}$ & RMSE & Change \\
    \hline
    \textsc{LatentPDE-MAP} & $0.01$ & $0.1223$ & -- \\
                         & $0$ & $0.1027$ & $-16.1\%$ \\
    \midrule \addlinespace  
    \textsc{LatentPDE-Enc} & $0.01$ & $0.1033$ & -- \\
                         & $0$ & $0.1347$ & $+30.4\%$ \\
    \hline
    \end{tabular}
    \caption{Final-refinement ablation. Setting $\lambda_{\mathrm{ref}}=0$ denotes disabling the post-diffusion refinement step.}
    \label{tab:ablation_refinement}
\end{table}

\paragraph{No source term.}
In our final ablation experiment, we investigate whether solving for the homogeneous case for our three PDE families has an impact on the results found in Table~\ref{tab:methods_comparison_source_on}. Results for this study remain consistent with those found in the inhomogeneous case. Please refer to Table~\ref{tab:methods_comparison_source_off} for full results.


\begin{table*}[h!]
\centering
\caption{Comparison of different models across PDE families with no source term (homogeneous case). As with Table \ref{tab:methods_comparison_source_on}, we report the better result between the MAP and encoder initial priors for 3D-Var and EnKF, chosen by the lowest aggregate RMSE in each setting. MAE is reported for deterministic point estimates, and CRPS for probabilistic predictions. A lower value is more desirable for all metrics listed.}
\label{tab:methods_comparison_source_off}
\resizebox{\textwidth}{!}{
\begin{tabular}{l cccc cccc cccc}
\toprule
 & \multicolumn{4}{c}{Advection--Diffusion} & \multicolumn{4}{c}{Klein--Gordon} & \multicolumn{4}{c}{Helmholtz} \\
\cmidrule(lr){2-5} \cmidrule(lr){6-9} \cmidrule(lr){10-13}
Method & RMSE & PSD & MAE & CRPS & RMSE & PSD & MAE & CRPS & RMSE & PSD & MAE & CRPS \\
\midrule
LatentPDE-MAP (Ours) & \textbf{0.080} & \textbf{1.049} & -- & \textbf{0.058} & \textbf{0.626} & 3.932 & -- & \textbf{0.415} & 0.023 & 2.918 & -- & 0.020 \\
LatentPDE-Enc (Ours) & 0.082 & 1.092 & -- & 0.062 & 0.653 & 3.982 & -- & 0.453 & \textbf{0.022} & \textbf{2.893} & -- & \textbf{0.018} \\
\midrule \addlinespace
EnKF & 0.082 & 12.500 & -- & 0.064 & 0.631 & 3.709 & -- & 0.491 & 0.039 & 7.467 & -- & 0.025 \\
3D-Var & 0.131 & 9.012 & 0.069 & -- & 0.628 & \textbf{3.413} & 0.496 & -- & 0.031 & 4.310 & 0.029 & -- \\
PINN & 0.169 & 8.224 & 0.136 & -- & 0.743 & 6.221 & 0.585 & -- & 0.080 & 5.500 & 0.067 & -- \\
\bottomrule
\end{tabular}
}
\end{table*}
\clearpage
\section{Computational Efficiency Analysis} \label{sec:appendix_compute}
For a single test sample on the advection--diffusion equation with source term setting, we show the inference times and associated number of parameters for all models used in Table \ref{tab:model_runtime} run on a single A5000 GPU. Because 3D-Var and EnKF are traditional data assimilation frameworks rather than learned models, they lack tunable weights in the deep learning sense. As seen, LatentPDE achieves up to a 64\% speedup when compared against FunDPS, likely due to the choice in latent parametrization and number of parameters. These runtimes are generated from a single reconstruction for a $128 \times 128$ resolution field with $s=5\%$ and $\sigma=0.15$.

\begin{table}[h!]
  \centering
  \caption{Comparison of model size and inference compute time.}
  \label{tab:model_runtime}
  \begin{tabular}{lrr}
    \toprule
    \textbf{Model} & \textbf{Parameters} & \textbf{Inference Time (s)} \\
    \midrule
    LatentPDE-MAP (Ours)  &  7.06M &  78.65 \\
    LatentPDE-Enc (Ours)  &  9.62M &  33.12 \\
    FunDPS & 184.6M  &   91.21 \\
    EnKF (MAP) & $-$ &  3.15 \\
    EnKF (Enc) & $-$ & 0.23 \\ 
    3D-Var (MAP) & $-$ &  3.84 \\
    3D-Var (Enc) & $-$ & 0.92 \\
    PINN &  58757 &  1.68 \\
    \bottomrule
  \end{tabular}
\end{table}




\end{document}